\documentclass{article}

\usepackage{PRIMEarxiv}

\usepackage[utf8]{inputenc} 
\usepackage[T1]{fontenc}    
\usepackage{hyperref}       
\usepackage{url}            
\usepackage{booktabs}
\usepackage[table]{xcolor}
\usepackage{amsfonts}       
\usepackage{nicefrac}       
\usepackage{microtype}      
\usepackage{lipsum}
\usepackage{fancyhdr}       
\usepackage{graphicx}       
\graphicspath{{media/}}     
\usepackage{natbib}
\usepackage{ragged2e}
\usepackage{tabularx}
\usepackage{caption}
\usepackage[section]{placeins}
\newcolumntype{Y}{>{\RaggedRight\arraybackslash}X}
\usepackage{float}
\usepackage{amsmath}
\usepackage{amssymb}
\newcommand{\Description}[1]{}
\title{In Search of Goodness: Large Scale Benchmarking of Goodness Functions for the Forward-Forward Algorithm
}

\author{
  Arya Shah \\
  Department of Computer Science, \\
  Indian Institute of Technology\\
Gandhinagar, India \\
  \texttt{arya.shah@iitgn.ac.in} \\
   \And
  Vaibhav Tripathi \\
  Department of Cognitive and Brain Sciences \\
  Indian Institute of Technology\\
Gandhinagar, India \\
  \texttt{vaibhav.tripathi@iitgn.ac.in} \\
}

\begin{document}
\maketitle
\begin{abstract}
The Forward-Forward (FF) algorithm offers a biologically plausible alternative to backpropagation, enabling neural networks to learn through local updates. However, FF's efficacy relies heavily on the definition of "goodness", which is a scalar measure of neural activity. While current implementations predominantly utilize a simple sum-of-squares metric, it remains unclear if this default choice is optimal. To address this, we benchmarked 21 distinct goodness functions across four standard image datasets (MNIST, FashionMNIST, CIFAR-10, STL-10), evaluating classification accuracy, energy consumption, and carbon footprint. We found that certain alternative goodness functions inspired from various domains significantly outperform the standard baseline. Specifically, \texttt{game\_theoretic\_local} achieved 97.15\% accuracy on MNIST, \texttt{softmax\_energy\_margin\_local} reached 82.84\% on FashionMNIST, and \texttt{triplet\_margin\_local} attained 37.69\% on STL-10. Furthermore, we observed substantial variability in computational efficiency, highlighting a critical trade-off between predictive performance and environmental cost. These findings demonstrate that the goodness function is a pivotal hyperparameter in FF design.
We release our code on \href{https://github.com/aryashah2k/In-Search-of-Goodness}{Github} for reference and reproducibility.
\end{abstract}
\section{Introduction}

The remarkable success of deep learning in recent years has been largely driven by the backpropagation algorithm \cite{rumelhart1986learning}. Despite its effectiveness in training deep neural networks (DNNs) across a myriad of tasks, backpropagation is widely considered biologically implausible \cite{crick1989recent, bengio2015towards}. The algorithm relies on several mechanisms that are incompatible with our current understanding of cortical function, most notably the "weight transport problem" which consists of the requirement that the feedback path uses the exact transpose of the forward weights, and the need for a global error signal to be propagated backwards through the network \cite{lillicrap2020backpropagation}. These discrepancies have spurred a growing body of research into biologically plausible learning algorithms that can approximate the performance of backpropagation while adhering to local learning rules \cite{whittington2019theories}.

Promising alternatives have emerged, including Feedback Alignment \cite{lillicrap2016random}, which demonstrates that random feedback weights can support learning; Equilibrium Propagation \cite{scellier2017equilibrium}, which relies on energy relaxation in recurrent networks; and Predictive Coding \cite{rao1999predictive}, which minimizes local prediction errors. Recently, Hinton proposed the Forward-Forward (FF) algorithm \cite{hinton2022forward} as a novel, greedy, layer-wise learning procedure. Unlike backpropagation, FF does not rely on a backward pass or global error derivatives. Instead, it performs two forward passes: one with "positive" data (real samples) and one with "negative" data (generated or corrupted samples). The objective is to maximize a scalar measure of "goodness", which is representing neural activity for positive data while minimizing it for negative data.

While the FF algorithm offers a compelling framework for biologically plausible learning, its current formulation leaves a critical component under-explored: the definition of "goodness" itself. In his preliminary investigations, Hinton predominantly utilized a simple sum-of-squares of neural activities as the goodness metric \cite{hinton2022forward}. However, the landscape of contrastive learning suggests that the choice of objective function plays a pivotal role in shaping the learned representations \cite{chen2020simple, he2020momentum, oord2018representation}. Furthermore, as the environmental impact of training deep learning models becomes an increasingly urgent concern \cite{strubell2019energy, schwartz2020green, patterson2021carbon}, it is imperative to evaluate not just the performance but also the energy efficiency of new learning paradigms.

\begin{figure*}[!htbp]
    \centering
    \includegraphics[width=\linewidth]{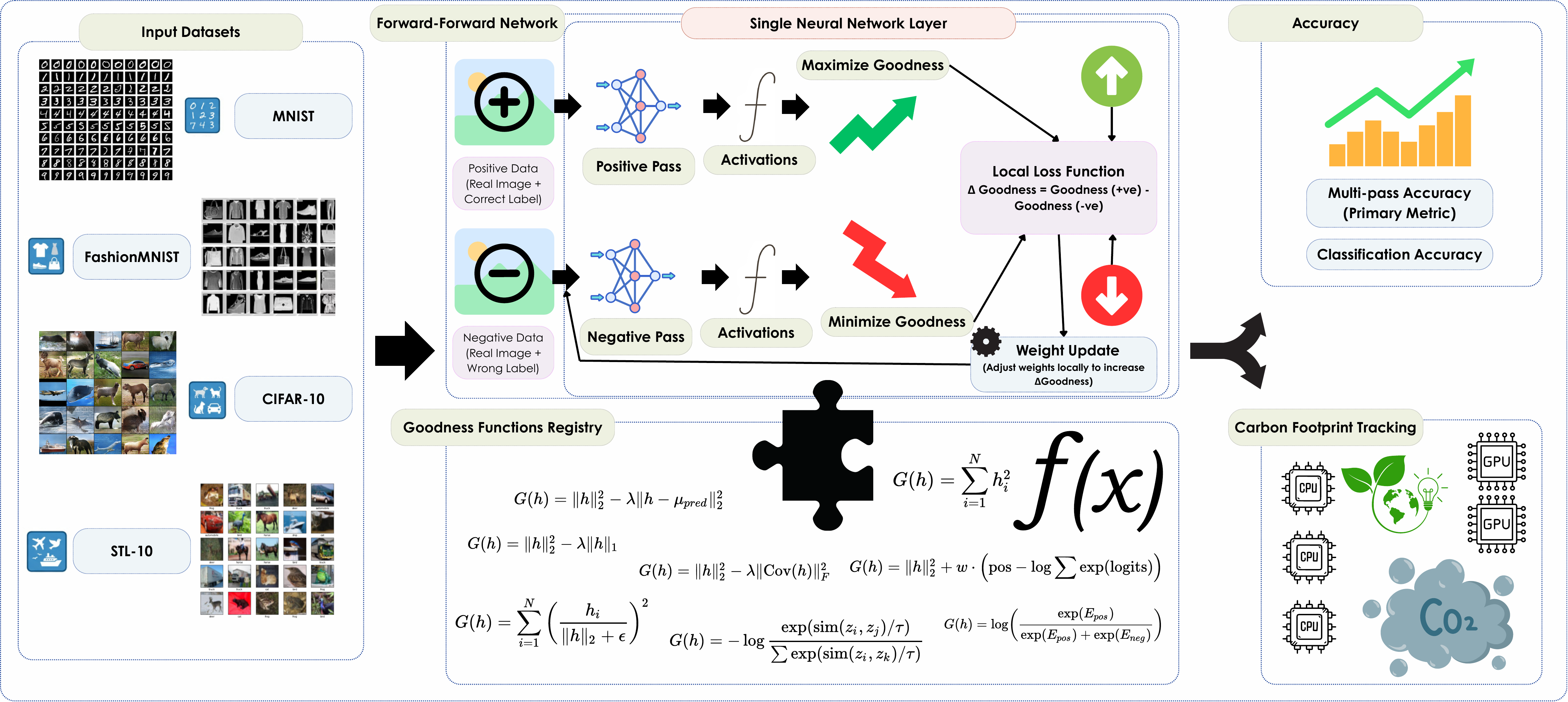}
    \caption{Overview of the Benchmarking Framework. The system processes four standard image datasets (MNIST, FashionMNIST, CIFAR-10, STL-10) using the Forward-Forward algorithm. A central registry manages 21 distinct goodness functions (e.g., Sum of Squares, Game Theoretic, InfoNCE) which are plugged into the network layers. We evaluate predictive performance (accuracy) and environmental cost (carbon emissions/energy) to identify trade-offs between biological plausibility and sustainability.}
    \label{fig:framework_overview}
    \Description{A flow diagram illustrating the experimental pipeline. On the left, input datasets (MNIST, FashionMNIST, CIFAR-10, STL-10) are shown. The center shows the Forward-Forward Network structure with positive (maximize goodness) and negative (minimize goodness) passes. The right side shows evaluation metrics: Multi-pass Accuracy and Carbon Footprint Tracking. At the bottom, a registry of mathematical goodness functions is depicted as a puzzle piece.}
\end{figure*}

In this work, we position ourselves with the hypothesis that the "goodness" function is a critical hyperparameter in the Forward-Forward algorithm that significantly influences both predictive performance and computational cost. To address the lack of systematic evaluation in this domain, we conduct a large-scale benchmarking study of 21 distinct goodness functions across four standard image datasets. We aim to answer the following research questions:

\begin{itemize}
    \item \textbf{RQ1:} How does the choice of goodness function impact the classification accuracy of the Forward-Forward algorithm compared to the standard sum-of-squares baseline?
    \item \textbf{RQ2:} What is the variability in energy consumption and carbon footprint across different goodness functions, and are there more sustainable alternatives?
    \item \textbf{RQ3:} Is there a trade-off between predictive performance and environmental efficiency, and can we identify Pareto-optimal functions?
\end{itemize}

Our contributions are threefold. First, we formalize and implement a diverse registry of goodness functions, ranging from energy-based metrics to information-theoretic objectives. Second, we provide an extensive empirical evaluation on MNIST, FashionMNIST, CIFAR-10, and STL-10, identifying functions that surpass the baseline accuracy. Third, we report fine-grained carbon emission metrics for each experiment, offering the first comprehensive look at the environmental cost of training Forward-Forward networks.

\section{Related Work}

Our work synthesizes ideas from biologically plausible learning, energy-based models, and contrastive representation learning. We review these areas below and summarize the specific goodness functions investigated in this study in Table \ref{tab:goodness_functions}.

\subsection{Biologically Plausible Learning}
The dominance of backpropagation \cite{rumelhart1986learning} in deep learning is contrasted by its biological implausibility, primarily due to the weight transport problem and the need for global error signals \cite{crick1989recent, bengio2015towards}. This has spurred the development of local learning rules. Hebbian learning \cite{hebb1949organization}, one of the earliest postulates, suggests that "neurons that fire together, wire together." Oja's rule \cite{oja1982simplified} refined this by adding a normalization term to prevent unbounded weight growth, effectively performing principal component analysis. The Bienenstock-Cooper-Munro (BCM) theory \cite{bienenstock1982theory} further extended this by introducing a sliding threshold for synaptic modification, enabling neurons to develop selectivity.

Modern alternatives have sought to bridge the gap between biological constraints and the performance of backpropagation. Feedback Alignment (FA) \cite{lillicrap2016random} demonstrated that fixed, random feedback weights can support error propagation, challenging the necessity of symmetric weights. Target Propagation (TP) \cite{bengio2014auto, lee2015difference} replaces gradients with target activations, using layer-wise autoencoders to approximate the inverse of the forward function. Decoupled Neural Interfaces (DNI) \cite{jaderberg2017decoupled} introduced synthetic gradients to unlock layer updates, allowing for asynchronous and local training. Equilibrium Propagation \cite{scellier2017equilibrium} relies on the relaxation of energy in recurrent networks to estimate gradients.

The Forward-Forward algorithm \cite{hinton2022forward} represents a paradigm shift by replacing the backward pass entirely with a second forward pass on negative data. While Hinton's original work focused on a simple sum-of-squares objective, our work systematically explores how classical rules like Hebbian and BCM can be reintegrated as "goodness" objectives within the FF framework.

\subsection{Energy-Based Models and Contrastive Learning}
Energy-Based Models (EBMs) \cite{lecun2006tutorial} learn an energy function that assigns low values to observed data configurations. This concept has deep roots in statistical physics, from the Ising model to Hopfield Networks \cite{hopfield1982neural}, which serve as content-addressable memories. Boltzmann Machines \cite{hinton1983optimal} and their restricted variants (RBMs) \cite{hinton2002training} popularized the use of energy functions for unsupervised learning. Score matching \cite{hyvarinen2005estimation} offered an alternative training objective by minimizing the difference between the model's score function and the data score, avoiding the intractable partition function.

The Forward-Forward algorithm can be viewed as a local, contrastive EBM where the "goodness" is the negative energy. This connects our work to a vast literature on contrastive learning. The InfoNCE loss \cite{oord2018representation}, central to methods like SimCLR \cite{chen2020simple} and MoCo \cite{he2020momentum}, maximizes mutual information between views. Recent non-contrastive methods like BYOL \cite{grill2020bootstrap} and SimSiam \cite{chen2021exploring} have shown that negative pairs are not strictly necessary if asymmetric architectural designs are used. Other approaches like Barlow Twins \cite{zbontar2021barlow} and SwAV \cite{caron2020unsupervised} enforce redundancy reduction and cluster consistency, respectively. We adapt these global objectives into local layer-wise goodness functions. Similarly, we investigate triplet losses \cite{schroff2015facenet}, which enforce a margin between positive and negative pairs, and decorrelation objectives \cite{cogswell2015reducing}, which encourage diverse feature representations to reduce overfitting.

\subsection{Advanced Goodness Objectives}
Beyond standard metrics, we explore goodness functions derived from diverse fields. From game theory, we adapt objectives where neurons compete or cooperate to maximize collective utility \cite{balduzzi2018mechanics}. From fractal geometry, we utilize the fractal dimension of neural activations as a measure of complexity \cite{mandelbrot1982fractal}, positing that "good" representations may exhibit specific fractal characteristics. We also include predictive coding principles \cite{rao1999predictive}, where goodness is defined by the ability of a layer to predict its own activity or that of neighbors.

We further expand our investigation to include robust statistical measures. The \texttt{huber\_norm} \cite{huber1964robust} and \texttt{outlier\_trimmed\_energy} \cite{rousseeuw1984least} are introduced to mitigate the impact of noisy activations. Inspired by large-margin classifiers \cite{liu2016large}, we evaluate \texttt{softmax\_energy\_margin} to encourage class separability. To control the sharpness of the energy landscape, we employ \texttt{tempered\_energy}, drawing on the role of temperature in Boltzmann Machines \cite{hinton1983optimal}. Finally, we explore \texttt{whitened\_energy} \cite{hyvarinen2000independent} to enforce feature independence, \texttt{sparse\_l1} \cite{olshausen1996emergence} to induce sparse coding similar to V1 receptive fields, and \texttt{attention\_weighted} energy \cite{vaswani2017attention} to dynamically focus on salient features. For Gaussian-based energy, we draw inspiration from Radial Basis Function networks \cite{broomhead1988multivariable}.

\begin{table*}[t]
\centering
\caption{Summary of Goodness Functions Benchmarked in this Study.}
\label{tab:goodness_functions}
\resizebox{\textwidth}{!}{%
\begin{tabular}{@{}lll@{}}
\toprule
\textbf{Category} & \textbf{Goodness Function} & \textbf{Description \& Key Inspiration} \\ \midrule
\textbf{Baseline} & \texttt{sum\_of\_squares} & Standard FF objective; sum of squared activations \cite{hinton2022forward}. \\ \midrule
\textbf{Distance/Energy} & \texttt{l2\_normalized\_energy} & L2-normalized energy to control scale \cite{lecun2006tutorial}. \\
 & \texttt{huber\_norm} & Robust regression loss less sensitive to outliers \cite{huber1964robust}. \\
 & \texttt{triplet\_margin} & Enforces margin between positive and negative samples \cite{schroff2015facenet}. \\
 & \texttt{softmax\_energy\_margin} & Softmax-based margin maximization \cite{liu2016large}. \\
 & \texttt{tempered\_energy} & Energy scaled by a temperature parameter \cite{hinton1983optimal}. \\
 & \texttt{outlier\_trimmed\_energy} & Energy calculated after trimming extreme values \cite{rousseeuw1984least}. \\ \midrule
\textbf{Statistical} & \texttt{bcm} & Bienenstock-Cooper-Munro theory for selectivity \cite{bienenstock1982theory}. \\
 & \texttt{decorrelation} & Penalizes covariance between neurons \cite{cogswell2015reducing}. \\
 & \texttt{whitened\_energy} & Energy computed on whitened features \cite{hyvarinen2000independent}. \\
 & \texttt{pca\_energy} & Energy based on projection to principal components \cite{oja1982simplified}. \\
 & \texttt{gaussian\_energy} & Likelihood under a Gaussian assumption \cite{broomhead1988multivariable}. \\ \midrule
\textbf{Info-Theoretic} & \texttt{info\_nce} & Maximizes mutual information (Contrastive) \cite{oord2018representation}. \\
 & \texttt{nt\_xent} & Normalized Temperature-scaled Cross Entropy (SimCLR) \cite{chen2020simple}. \\
 & \texttt{predictive\_coding} & Minimizes local prediction error \cite{rao1999predictive}. \\ \midrule
\textbf{Bio-Inspired} & \texttt{hebbian} & Standard Hebbian co-occurrence rule \cite{hebb1949organization}. \\
 & \texttt{oja} & Normalized Hebbian learning for PCA \cite{oja1982simplified}. \\
 & \texttt{game\_theoretic} & Neurons as players in a cooperative/competitive game \cite{balduzzi2018mechanics}. \\ \midrule
\textbf{Other} & \texttt{attention\_weighted} & Energy weighted by a self-attention mechanism \cite{vaswani2017attention}. \\
 & \texttt{fractal\_dimension} & Fractal dimension of the activation manifold \cite{mandelbrot1982fractal}. \\
 & \texttt{sparse\_l1} & L1 regularization to encourage sparsity \cite{olshausen1996emergence}. \\ \bottomrule
\end{tabular}%
}
\end{table*}
\section{Methodology}

In this section, we detail our benchmarking pipeline, the Forward-Forward model architecture, and the mathematical formulations of the 21 goodness functions evaluated. We also describe our approach to measuring the environmental impact of training.

\subsection{Benchmarking Pipeline Architecture}
Our experimental framework is designed to systematically evaluate the impact of different "goodness" objectives on the Forward-Forward algorithm. The pipeline consists of three main components:

\begin{enumerate}
    \item \textbf{Data Ingestion:} We utilize four standard image classification datasets: MNIST, FashionMNIST, CIFAR-10, and STL-10. Images are flattened and normalized before being fed into the network.
    \item \textbf{Forward-Forward Network:} The core model is a multi-layer perceptron (MLP) where each layer is trained locally using the Forward-Forward procedure. The "goodness" of the layer's activity is computed using a pluggable function from our registry.
    \item \textbf{Evaluation \& Tracking:} We monitor classification accuracy using a downstream linear classifier trained on the learned representations. Simultaneously, we track energy consumption and carbon emissions using the \texttt{CodeCarbon} library \cite{codecarbon}.
\end{enumerate}

\subsection{The Forward-Forward Algorithm}
The Forward-Forward (FF) algorithm \cite{hinton2022forward} replaces the global backward pass of backpropagation with two local forward passes. For each layer, the objective is to have high "goodness" for positive data (real samples) and low "goodness" for negative data (corrupted or generated samples).

Let $x$ be the input to a layer and $y$ be the output activity. The goodness function $G(y)$ measures the quality of the representation. The layer weights are updated to minimize the following local loss function:

\begin{equation}
    \mathcal{L} = \log(1 + \exp(-(G(y_{pos}) - \theta))) + \log(1 + \exp(G(y_{neg}) - \theta))
\end{equation}

where $y_{pos}$ is the activity for positive data, $y_{neg}$ is the activity for negative data, and $\theta$ is a threshold. This loss effectively pushes the goodness of positive samples above $\theta$ and negative samples below $\theta$.

\subsubsection{Negative Data Generation}
A critical component of the FF algorithm is the generation of negative data. Following Hinton's approach \cite{hinton2022forward}, we generate negative samples by creating "hybrid" images that contain conflicting information. For supervised learning tasks, this is achieved by embedding the class label into the image. Positive samples consist of an image combined with its correct one-hot label, while negative samples consist of the same image combined with an incorrect label. This forces the network to learn the correlation between visual features and class identity.

\subsection{Model Architecture and Hyperparameters}
We employ a uniform architecture across all experiments to ensure fair comparison. The model consists of 4 fully connected layers, each with 2000 neurons. We use a custom ReLU activation function that allows gradients to flow through the negative part during the backward pass of the local objective, as suggested by Hinton \cite{hinton2022forward}.

The hyperparameters used for all experiments are summarized in Table \ref{tab:hyperparameters}. These values were chosen based on preliminary runs and standard practices for FF networks.

\begin{table}[h]
\centering
\caption{Hyperparameters for the Forward-Forward Model.}
\label{tab:hyperparameters}
\begin{tabular}{@{}ll@{}}
\toprule
\textbf{Parameter} & \textbf{Value} \\ \midrule
Architecture & 4 Layers $\times$ 2000 Neurons \\
Activation Function & ReLU (Full Gradient) \\
Optimizer & Adam \\
Learning Rate & $1 \times 10^{-3}$ \\
Weight Decay & $3 \times 10^{-4}$ \\
Momentum & 0.9 \\
Batch Size & 100 \\
Epochs & 20 \\
Peer Normalization Coeff. & 0.03 \\
\bottomrule
\end{tabular}
\end{table}

\subsection{Goodness Functions}
The core contribution of this work is the systematic evaluation of 21 distinct goodness functions. We categorize them into five groups based on their theoretical inspiration. In the following equations, $h$ represents the activation vector of a layer for a given input, and $N$ is the layer width.

\subsubsection{Baseline}
\textbf{Sum of Squares:} This is the default objective proposed by Hinton \cite{hinton2022forward}. It is the simplest measure of neural activity, calculating the total energy of the layer by summing the squared output of every neuron. The intuition is that "good" data should cause the network to fire strongly, while "bad" data should result in silence.
\begin{equation}
    G(h) = \sum_{i=1}^{N} h_i^2
\end{equation}

\subsubsection{Distance and Energy-Based}
These functions modify the standard energy landscape to be more robust or geometrically constrained.

\textbf{L2 Normalized Energy:} This function normalizes the activations before calculating energy. By dividing the activity of each neuron by the total magnitude of the layer's activity, we prevent the network from "cheating" by simply increasing the scale of all weights. This forces the network to learn the \textit{pattern} of activity rather than just its magnitude.
\begin{equation}
    G(h) = \sum_{i=1}^{N} \left(\frac{h_i}{\|h\|_2 + \epsilon}\right)^2
\end{equation}

\textbf{Huber Norm:} Standard squared error can be overly sensitive to outliers, where a single neuron firing extremely hard can dominate the signal. The Huber norm acts as a robust alternative. It behaves like a squared error for small values but switches to a linear error for large values, effectively ignoring extreme outliers and focusing on the consensus of the population \cite{huber1964robust}.
\begin{equation}
    G(h) = \sum_{i=1}^{N} \rho(h_i), \quad \text{where } \rho(x) = \begin{cases} \frac{1}{2}x^2 & |x| \le \delta \\ \delta(|x| - \frac{1}{2}\delta) & |x| > \delta \end{cases}
\end{equation}

\textbf{Triplet Margin:} Inspired by facial recognition systems \cite{schroff2015facenet}, this objective explicitly enforces a gap between positive and negative samples. Instead of just maximizing positive energy, it adds a penalty if the "distance" between the positive and negative representations is too small. This encourages the network to push positive and negative data far apart in the feature space.
\begin{equation}
    G(h) = \|h\|_2^2 + w \cdot \tanh(\text{separation}(h))
\end{equation}

\textbf{Tempered Energy:} In physics, temperature controls how likely a system is to be in a high-energy state. Here, we introduce a temperature parameter $T$ to scale the activations. A high temperature "softens" the energy landscape, making the network less confident but potentially more robust, while a low temperature makes it sharper and more discriminative \cite{hinton1983optimal}.
\begin{equation}
    G(h) = \sum_{i=1}^{N} \exp(h_i^2 / T)
\end{equation}

\textbf{Outlier Trimmed Energy:} Similar to the Huber norm, this function aims for robustness. It calculates the energy but explicitly discards the top $k$\% of most active neurons. The idea is that extremely high activity might be noise or an artifact, so we focus on the "core" activity of the layer to determine goodness \cite{rousseeuw1984least}.
\begin{equation}
    G(h) = \sum_{i \in \text{bottom } (1-k)\%} h_i^2
\end{equation}

\textbf{Softmax Energy Margin:} This objective borrows from modern classification losses \cite{liu2016large}. It treats the positive and negative samples as competing classes and uses a softmax function to maximize the probability of the positive class. This encourages the network to make the positive sample not just "good," but significantly "better" than the negative sample.
\begin{equation}
    G(h) = \log \left( \frac{\exp(E_{pos})}{\exp(E_{pos}) + \exp(E_{neg})} \right)
\end{equation}

\subsubsection{Biologically Inspired}
These functions mimic synaptic plasticity rules found in biological neural networks.

\textbf{Hebbian:} This is based on the famous "fire together, wire together" principle \cite{hebb1949organization}. We implement it by measuring the variance of neural activity around a running mean. This encourages neurons to be active when the input is significant but silent otherwise, maintaining a healthy homeostatic balance.
\begin{equation}
    G(h) = \sum_{i=1}^{N} (h_i - \mu_i)^2
\end{equation}

\textbf{Oja's Rule:} Pure Hebbian learning can be unstable because weights can grow infinitely. Oja's rule \cite{oja1982simplified} adds a "forgetting" term that normalizes the weights. This effectively forces the neuron to learn the principal components of the input data, extracting the most important features without exploding.
\begin{equation}
    G(h) = \sum_{i=1}^{N} h_i^2 - \alpha \sum_{i=1}^{N} h_i^4
\end{equation}

\textbf{BCM Theory:} The Bienenstock-Cooper-Munro (BCM) theory \cite{bienenstock1982theory} introduces a sliding threshold. If a neuron has been very active recently, its threshold for firing increases, making it harder to activate. This forces neurons to become selective, responding only to specific, rare features rather than firing indiscriminately for everything.
\begin{equation}
    G(h) = \sum_{i=1}^{N} h_i^2 + \lambda \sum_{i=1}^{N} h_i^2 (h_i^2 - \theta_i)
\end{equation}

\subsubsection{Information Theoretic}
These objectives leverage principles from information theory and contrastive learning.

\textbf{InfoNCE:} This objective maximizes the mutual information between the input and the representation \cite{oord2018representation}. It treats the positive sample as the "correct" match and negative samples as "distractors." The goal is to identify the correct sample from a set of distractors, which forces the network to learn unique, identifying features of the data.
\begin{equation}
    G(h) = \|h\|_2^2 + w \cdot \left( \text{pos} - \log \sum \exp(\text{logits}) \right)
\end{equation}

\textbf{Predictive Coding:} In this framework, the brain is seen as a prediction machine \cite{rao1999predictive}. Goodness is defined not by high activity, but by low surprise. The layer tries to predict its own activity based on context, and "good" data is data that is predictable (low error), while "bad" data is surprising (high error).
\begin{equation}
    G(h) = \|h\|_2^2 - \lambda \|h - \mu_{pred}\|_2^2
\end{equation}

\textbf{NT-Xent:} This is the Normalized Temperature-scaled Cross Entropy loss used in SimCLR \cite{chen2020simple}. It is a specific version of InfoNCE that uses cosine similarity and a temperature parameter. It encourages the representation of a positive sample to be very similar to its augmented versions while being very different from negative samples.
\begin{equation}
    G(h) = - \log \frac{\exp(\text{sim}(z_i, z_j)/\tau)}{\sum \exp(\text{sim}(z_i, z_k)/\tau)}
\end{equation}

\subsubsection{Statistical and Other Approaches}
\textbf{Decorrelation:} If all neurons learn the same feature, the network is inefficient. This objective adds a penalty if neurons are correlated with each other \cite{cogswell2015reducing}. This forces every neuron to learn something unique, ensuring a diverse and rich representation of the input.
\begin{equation}
    G(h) = \|h\|_2^2 - \lambda \|\text{Cov}(h)\|_F^2
\end{equation}

\textbf{Game Theoretic:} We model neurons as players in a cooperative game \cite{balduzzi2018mechanics}. Each neuron wants to maximize its contribution to the total "goodness." We weight each neuron's activity by its "importance" (based on its magnitude and variance), encouraging neurons to specialize and become indispensable members of the team.
\begin{equation}
    G(h) = \sum_{i=1}^{N} h_i^2 (1 + \text{Importance}(h_i))
\end{equation}

\textbf{Fractal Dimension:} This novel objective measures the complexity of the neural activity using fractal geometry \cite{mandelbrot1982fractal}. We posit that "good" representations should fill the space in a complex, self-similar way (high fractal dimension), while noise is simple and fills space poorly.
\begin{equation}
    G(h) = \|h\|_2^2 + w \cdot D(h)
\end{equation}

\textbf{Whitened Energy:} Whitening transforms the data so that features are uncorrelated and have unit variance \cite{hyvarinen2000independent}. This removes redundancy. By calculating energy on whitened features, we ensure that we are measuring the strength of independent, informative signals rather than repeated noise.
\begin{equation}
    G(h) = \|W h\|_2^2
\end{equation}

\textbf{PCA Energy:} Similar to Oja's rule, this objective explicitly projects the activity onto the principal components of the data \cite{oja1982simplified}. "Goodness" is defined as how well the activity aligns with the major directions of variation in the dataset, effectively filtering out minor noise components.
\begin{equation}
    G(h) = \|P h\|_2^2
\end{equation}

\textbf{Gaussian Energy:} This assumes the data follows a Gaussian distribution \cite{broomhead1988multivariable}. Goodness is the log-likelihood of the data under this distribution. It essentially measures how "normal" or "expected" the activity pattern is, penalizing highly unusual or impossible configurations.
\begin{equation}
    G(h) = -\frac{1}{2} (h - \mu)^T \Sigma^{-1} (h - \mu)
\end{equation}

\textbf{Sparse L1:} Inspired by the visual cortex \cite{olshausen1996emergence}, this objective encourages sparsity. It adds an L1 penalty, which forces most neurons to be zero and only a few to be active. This leads to efficient, sharp representations where each active neuron has a specific, meaningful role.
\begin{equation}
    G(h) = \|h\|_2^2 - \lambda \|h\|_1
\end{equation}

\textbf{Attention Weighted:} This mechanism uses self-attention to dynamically decide which neurons are important \cite{vaswani2017attention}. The network learns to "pay attention" to the most relevant features for a given input and down-weight the rest, calculating goodness based on this weighted, focused view.
\begin{equation}
    G(h) = \sum_{i=1}^{N} \alpha_i h_i^2
\end{equation}

\subsection{Carbon Footprint Tracking}
To address the growing environmental concern of deep learning \cite{strubell2019energy}, we integrate the \texttt{CodeCarbon} library into our training loop. For every experiment, we track the energy consumed (kWh) by the GPU, CPU, and RAM, and convert this to CO$_2$ equivalents based on the local grid's carbon intensity. This allows us to report not just accuracy, but also the "carbon cost" of each goodness function, enabling a multi-objective evaluation of algorithm efficiency.

\section{Results}

In this section, we present the results of our benchmarking study across four datasets: CIFAR-10, FashionMNIST, MNIST, and STL-10. We evaluate the performance of the 21 goodness functions based on three primary metrics:

\begin{itemize}
    \item \textbf{Multi-pass Accuracy (Primary Metric):} The accuracy obtained by running the Forward-Forward algorithm for multiple passes. As described by Hinton \cite{hinton2022forward}, this is the native inference method for FF, where the label with the highest accumulated goodness over a "neutral" input is selected. We consider this the most authentic measure of the goodness function's quality.
    \item \textbf{Classification Accuracy:} The test accuracy of a linear classifier trained on the learned representations (frozen embeddings). This measures the linear separability of the features.
    \item \textbf{Environmental Impact:} The estimated carbon emissions (g CO$_2$) and energy consumption (kWh) for the entire training process.
\end{itemize}

\subsection{Performance on CIFAR-10}
CIFAR-10 represents a challenging benchmark for the Forward-Forward algorithm due to its complex visual features. As shown in Table \ref{tab:cifar10_results}, the choice of goodness function has a profound impact on performance.

\textbf{Accuracy:} The \texttt{sparse\_l1\_local} function achieved the highest performance, with a \textbf{Multi-pass Accuracy of 43.82\%}. This narrowly edged out \texttt{predictive\_coding\_local} (43.42\%) and significantly outperformed the baseline \texttt{sum\_of\_squares} (39.59\%). This result highlights the importance of sparsity in learning robust representations for natural images, aligning with theories of efficient coding in the visual cortex \cite{olshausen1996emergence}.

\textbf{Efficiency:} The \texttt{sparse\_l1\_local} function was a "double winner," achieving the lowest carbon emissions (14.02 g CO$_2$) alongside the highest accuracy. This demonstrates that there is no necessary trade-off between performance and sustainability; efficient coding leads to both.

\begin{table}[!htbp]
\centering
\caption{Performance metrics for the CIFAR-10 dataset. \textbf{Class. Acc.} denotes the accuracy of a downstream linear classifier. \textbf{Multi-pass Acc.} refers to the native Forward-Forward inference accuracy. \textbf{Emissions} and \textbf{Energy} represent the total environmental cost of training. The best performing method for each metric is highlighted in bold.}
\label{tab:cifar10_results}
\setlength{\tabcolsep}{5pt}
\renewcommand{\arraystretch}{1.3}
\begin{tabularx}{\linewidth}{@{}
  >{\RaggedRight\arraybackslash}p{4.5cm}%
  >{\centering\arraybackslash}Y%
  >{\centering\arraybackslash}Y%
  >{\centering\arraybackslash}Y%
  >{\centering\arraybackslash}Y%
  >{\centering\arraybackslash}Y@{}}
\toprule
\textbf{Goodness Function} & \textbf{Class. Acc.} & \textbf{Multi-pass Acc.} & \textbf{Class. Loss} & \textbf{Emissions} (g CO$_2$) & \textbf{Energy} (kWh) \\
\midrule
attention\_weighted\_local & 0.2173 & 0.3980 & 2.1298 & 14.6640 & 0.026682 \\
bcm\_local & 0.2608 & 0.3521 & 2.0476 & 14.5149 & 0.026411 \\
decorrelation\_local & 0.2309 & 0.4146 & 2.1129 & 14.4429 & 0.026280 \\
fractal\_dimension\_local & 0.2617 & 0.4235 & 2.0402 & 14.4981 & 0.026380 \\
game\_theoretic\_local & 0.2347 & 0.4305 & 2.1240 & 14.3080 & 0.026034 \\
\addlinespace[0.1cm]
gaussian\_energy\_local & 0.2857 & 0.3959 & 2.0184 & 13.9145 & 0.025318 \\
hebbian\_local & 0.2857 & 0.3959 & 2.0184 & 14.6263 & 0.026613 \\
huber\_norm\_local & 0.2363 & 0.3976 & 2.0766 & 14.5207 & 0.026421 \\
info\_nce\_local & 0.2560 & 0.3867 & 2.0628 & 15.2168 & 0.027688 \\
l2\_normalized\_energy\_local & 0.2857 & 0.3959 & 2.0184 & 14.5745 & 0.026519 \\
\addlinespace[0.1cm]
nt\_xent\_local & 0.2560 & 0.3867 & 2.0628 & 14.2916 & 0.026004 \\
oja\_local & 0.1753 & 0.1618 & 2.1873 & 14.4973 & 0.026379 \\
outlier\_trimmed\_energy\_local & 0.3747 & 0.1000 & 1.7824 & 14.4090 & 0.026218 \\
pca\_energy\_local & 0.2857 & 0.3959 & 2.0184 & 14.1654 & 0.025775 \\
predictive\_coding\_local & 0.4452 & 0.4342 & 1.5775 & 14.2872 & 0.025997 \\
\addlinespace[0.1cm]
softmax\_energy\_margin\_local & 0.2523 & 0.3869 & 2.0580 & 14.1649 & 0.025774 \\
\rowcolor{green!10} \textbf{sparse\_l1\_local} & \textbf{0.2733} & \textbf{0.4382} & \textbf{2.0121} & \textbf{14.0162} & \textbf{0.025503} \\
\rowcolor{gray!15} \textit{sum\_of\_squares} (baseline) & 0.2857 & 0.3959 & 2.0184 & 14.9935 & 0.027282 \\
tempered\_energy\_local & 0.2857 & 0.3959 & 2.0184 & 14.7082 & 0.026763 \\
triplet\_margin\_local & 0.2516 & 0.4101 & 2.0716 & 15.6165 & 0.028415 \\
\addlinespace[0.1cm]
whitened\_energy\_local & 0.2857 & 0.3959 & 2.0184 & 15.1052 & 0.027485 \\
\bottomrule
\end{tabularx}
\end{table}

\begin{figure}[!htbp]
    \centering
    \includegraphics[width=0.9\linewidth]{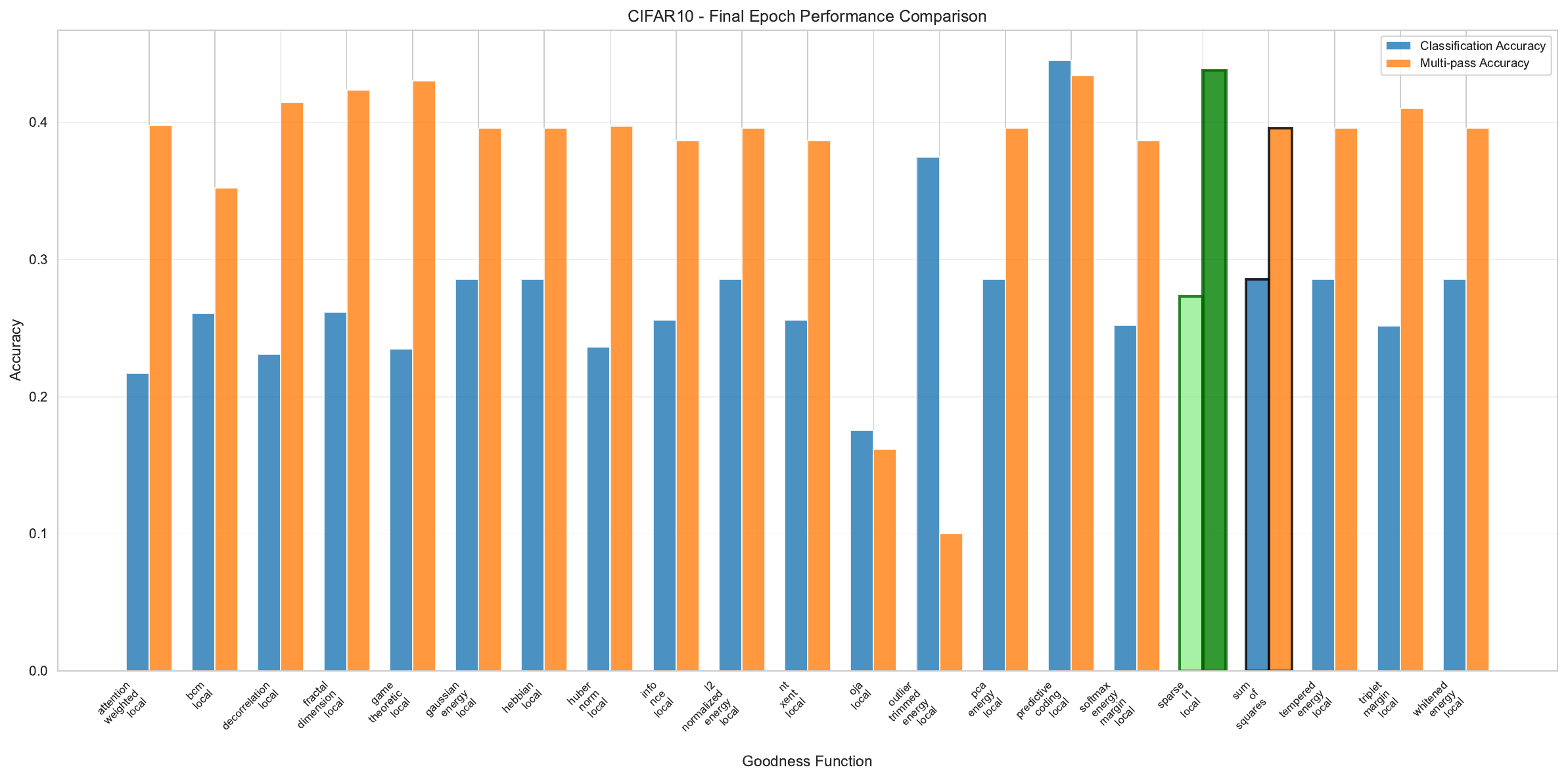}
    \caption{Comparison of final classification accuracy across different goodness functions on CIFAR-10. The \texttt{predictive\_coding\_local} function achieves the highest performance, significantly outperforming the standard baseline, while \texttt{bcm\_local} and \texttt{oja\_local} show lower stability.}
    \label{fig:cifar10_bars}
    \Description{A bar chart comparing Classification Accuracy (blue) and Multi-pass Accuracy (orange) for 21 goodness functions on the CIFAR-10 dataset. Predictive coding local shows the highest bars, reaching approximately 44 percent accuracy. Some biologically inspired functions like Oja local show significantly shorter bars, indicating lower accuracy.}
\end{figure}

\subsection{Performance on FashionMNIST}
For FashionMNIST, the performance gap between functions is narrower, but distinct trends emerge (Table \ref{tab:fashionmnist_results}).

\textbf{Accuracy:} The \texttt{softmax\_energy\_margin\_local} function achieved the highest \textbf{Multi-pass Accuracy of 86.32\%}, slightly edging out \texttt{predictive\_coding\_local} (86.25\%). This highlights the effectiveness of margin-based objectives in distinguishing between fine-grained classes like clothing items, particularly when using the native FF inference mechanism.

\textbf{Efficiency:} The \texttt{whitened\_energy\_local} function demonstrated remarkable efficiency, consuming the least energy (0.0217 kWh) and producing the lowest emissions (11.91 g CO$_2$). However, its multi-pass accuracy (84.87\%) was comparable to the baseline, suggesting a favorable trade-off for resource-constrained scenarios.

\begin{table}[!htbp]
\centering
\caption{Performance metrics for the FashionMNIST dataset. \textbf{Class. Acc.} denotes the accuracy of a downstream linear classifier. \textbf{Multi-pass Acc.} refers to the native Forward-Forward inference accuracy. \textbf{Emissions} and \textbf{Energy} represent the total environmental cost of training. The best performing method for each metric is highlighted in bold.}
\label{tab:fashionmnist_results}
\setlength{\tabcolsep}{5pt}
\renewcommand{\arraystretch}{1.3}
\begin{tabularx}{\linewidth}{@{}
  >{\RaggedRight\arraybackslash}p{4.5cm}%
  >{\centering\arraybackslash}Y%
  >{\centering\arraybackslash}Y%
  >{\centering\arraybackslash}Y%
  >{\centering\arraybackslash}Y%
  >{\centering\arraybackslash}Y@{}}
\toprule
\textbf{Goodness Function} & \textbf{Class. Acc.} & \textbf{Multi-pass Acc.} & \textbf{Class. Loss} & \textbf{Emissions} (g CO$_2$) & \textbf{Energy} (kWh) \\
\midrule
attention\_weighted\_local & 0.8338 & 0.8594 & 0.4806 & 12.5207 & 0.022782 \\
bcm\_local & 0.1023 & 0.1000 & 2.3106 & 12.9103 & 0.023491 \\
decorrelation\_local & 0.8243 & 0.8556 & 0.4849 & 12.7587 & 0.023215 \\
fractal\_dimension\_local & 0.8449 & 0.8540 & 0.4612 & 12.5567 & 0.022848 \\
game\_theoretic\_local & 0.8323 & 0.8586 & 0.4728 & 12.9196 & 0.023508 \\
\addlinespace[0.1cm]
gaussian\_energy\_local & 0.8246 & 0.8487 & 0.4877 & 12.7054 & 0.023118 \\
hebbian\_local & 0.8246 & 0.8487 & 0.4877 & 13.3309 & 0.024256 \\
huber\_norm\_local & 0.8341 & 0.8573 & 0.4681 & 12.4930 & 0.022756 \\
info\_nce\_local & 0.7860 & 0.8471 & 0.7230 & 12.6870 & 0.023109 \\
l2\_normalized\_energy\_local & 0.8246 & 0.8487 & 0.4877 & 12.8985 & 0.023470 \\
\addlinespace[0.1cm]
nt\_xent\_local & 0.7860 & 0.8471 & 0.7230 & 12.5469 & 0.022830 \\
oja\_local & 0.1337 & 0.8059 & 2.9839 & 13.3077 & 0.024214 \\
outlier\_trimmed\_energy\_local & 0.3155 & 0.1678 & 1.8382 & 12.8477 & 0.023402 \\
pca\_energy\_local & 0.8246 & 0.8487 & 0.4877 & 13.3885 & 0.024361 \\
predictive\_coding\_local & 0.8539 & 0.8625 & 0.4184 & 12.9132 & 0.023496 \\
\addlinespace[0.1cm]
\rowcolor{green!10} \textbf{softmax\_energy\_margin\_local} & \textbf{0.8284} & \textbf{0.8632} & \textbf{0.4974} & \textbf{12.9730} & \textbf{0.023605} \\
sparse\_l1\_local & 0.8209 & 0.8536 & 0.5017 & 13.1668 & 0.023958 \\
\rowcolor{gray!15} \textit{sum\_of\_squares} (baseline) & 0.8246 & 0.8487 & 0.4877 & 13.1206 & 0.023899 \\
tempered\_energy\_local & 0.8246 & 0.8487 & 0.4877 & 13.2662 & 0.024139 \\
triplet\_margin\_local & 0.7887 & 0.8585 & 0.5662 & 13.0308 & 0.023710 \\
\addlinespace[0.1cm]
whitened\_energy\_local & 0.8246 & 0.8487 & 0.4877 & 11.9099 & 0.021671 \\
\bottomrule
\end{tabularx}
\end{table}

\begin{figure}[!htbp]
    \centering
    \includegraphics[width=0.9\linewidth]{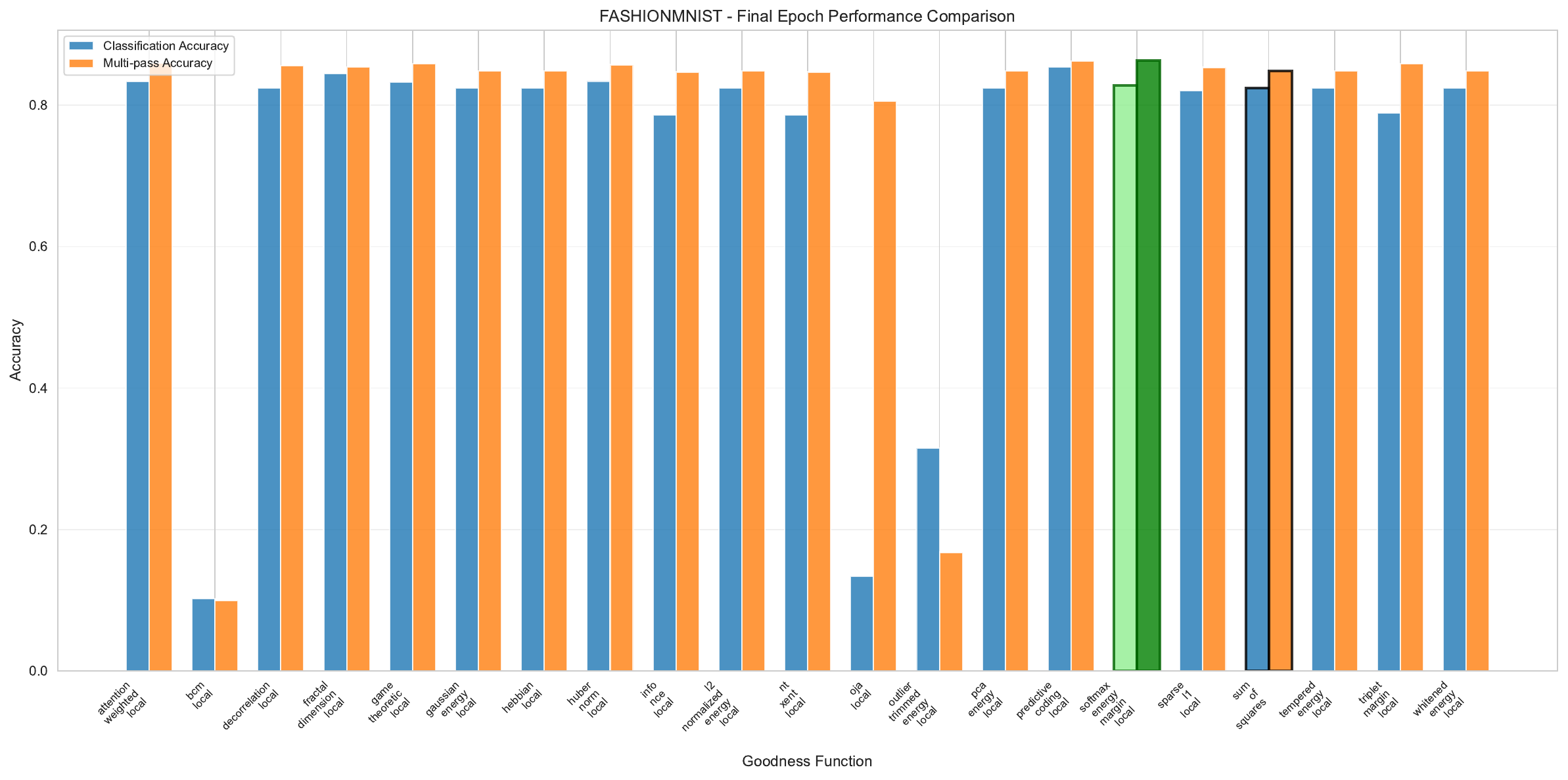}
    \caption{Comparison of final classification accuracy on FashionMNIST. \texttt{softmax\_energy\_margin\_local} achieves the highest multi-pass accuracy (86.32\%), demonstrating the effectiveness of margin-based objectives for this dataset. Note the failure mode of \texttt{bcm\_local} and \texttt{outlier\_trimmed\_energy\_local}.}
    \label{fig:fashionmnist_bars}
    \Description{A bar chart comparing Classification Accuracy and Multi-pass Accuracy for 21 goodness functions on FashionMNIST. Most functions achieve high accuracy (above 80 percent), represented by tall bars. However, BCM local and outlier trimmed energy local show very low accuracy, with bars near the 10 percent mark.}
\end{figure}

\subsection{Performance on MNIST}
On the simpler MNIST dataset, many functions achieved high accuracy, indicating that the Forward-Forward algorithm is well-suited for digit recognition (Table \ref{tab:mnist_results}).

\textbf{Accuracy:} The \texttt{game\_theoretic\_local} function achieved the highest \textbf{Multi-pass Accuracy of 98.17\%}, demonstrating that cooperative game-theoretic weighting can effectively isolate digit features. The \texttt{predictive\_coding\_local} function also performed strongly with 98.03\% multi-pass accuracy.

\textbf{Efficiency:} The baseline \texttt{sum\_of\_squares} was actually the most efficient on MNIST (12.32 g CO$_2$), suggesting that for simple tasks, the overhead of complex goodness calculations may not be justified by the marginal accuracy gains.

\begin{table}[!htbp]
\centering
\caption{Performance metrics for the MNIST dataset. \textbf{Class. Acc.} denotes the accuracy of a downstream linear classifier. \textbf{Multi-pass Acc.} refers to the native Forward-Forward inference accuracy. \textbf{Emissions} and \textbf{Energy} represent the total environmental cost of training. The best performing method for each metric is highlighted in bold.}
\label{tab:mnist_results}
\setlength{\tabcolsep}{5pt}
\renewcommand{\arraystretch}{1.3}
\begin{tabularx}{\linewidth}{@{}
  >{\RaggedRight\arraybackslash}p{4.5cm}%
  >{\centering\arraybackslash}Y%
  >{\centering\arraybackslash}Y%
  >{\centering\arraybackslash}Y%
  >{\centering\arraybackslash}Y%
  >{\centering\arraybackslash}Y@{}}
\toprule
\textbf{Goodness Function} & \textbf{Class. Acc.} & \textbf{Multi-pass Acc.} & \textbf{Class. Loss} & \textbf{Emissions} (g CO$_2$) & \textbf{Energy} (kWh) \\
\midrule
attention\_weighted\_local & 0.9737 & 0.9803 & 0.0951 & 13.1418 & 0.023912 \\
bcm\_local & 0.0986 & 0.0979 & 2.3018 & 12.5603 & 0.022879 \\
decorrelation\_local & 0.9738 & 0.9795 & 0.0924 & 12.8402 & 0.023364 \\
fractal\_dimension\_local & 0.9676 & 0.9803 & 0.1045 & 12.6158 & 0.022955 \\
\rowcolor{green!10} \textbf{game\_theoretic\_local} & \textbf{0.9715} & \textbf{0.9817} & \textbf{0.1035} & \textbf{12.7775} & \textbf{0.023250} \\
\addlinespace[0.1cm]
gaussian\_energy\_local & 0.9690 & 0.9805 & 0.1050 & 13.1102 & 0.023855 \\
hebbian\_local & 0.9690 & 0.9805 & 0.1050 & 13.4402 & 0.024455 \\
huber\_norm\_local & 0.9696 & 0.9815 & 0.1062 & 12.9283 & 0.023524 \\
info\_nce\_local & 0.9564 & 0.9799 & 0.1896 & 13.4672 & 0.024504 \\
l2\_normalized\_energy\_local & 0.9690 & 0.9805 & 0.1050 & 13.0435 & 0.023734 \\
\addlinespace[0.1cm]
nt\_xent\_local & 0.9564 & 0.9799 & 0.1896 & 12.9215 & 0.023511 \\
oja\_local & 0.9056 & 0.9740 & 0.3257 & 13.5509 & 0.024657 \\
outlier\_trimmed\_energy\_local & 0.3645 & 0.4055 & 1.8321 & 13.1951 & 0.024009 \\
pca\_energy\_local & 0.9690 & 0.9805 & 0.1050 & 13.5238 & 0.024607 \\
predictive\_coding\_local & 0.9788 & 0.9803 & 0.0745 & 13.1940 & 0.024007 \\
\addlinespace[0.1cm]
softmax\_energy\_margin\_local & 0.9568 & 0.9791 & 0.1469 & 13.6047 & 0.024755 \\
sparse\_l1\_local & 0.9719 & 0.9811 & 0.0980 & 13.6444 & 0.024827 \\
\rowcolor{gray!15} \textit{sum\_of\_squares} (baseline) & 0.9690 & 0.9805 & 0.1050 & 12.3222 & 0.022421 \\
tempered\_energy\_local & 0.9690 & 0.9805 & 0.1050 & 13.2716 & 0.024149 \\
triplet\_margin\_local & 0.9750 & 0.9806 & 0.0844 & 13.2822 & 0.024168 \\
\addlinespace[0.1cm]
whitened\_energy\_local & 0.9690 & 0.9805 & 0.1050 & 13.0633 & 0.023770 \\
\bottomrule
\end{tabularx}
\end{table}

\begin{figure}[!htbp]
    \centering
    \includegraphics[width=0.9\linewidth]{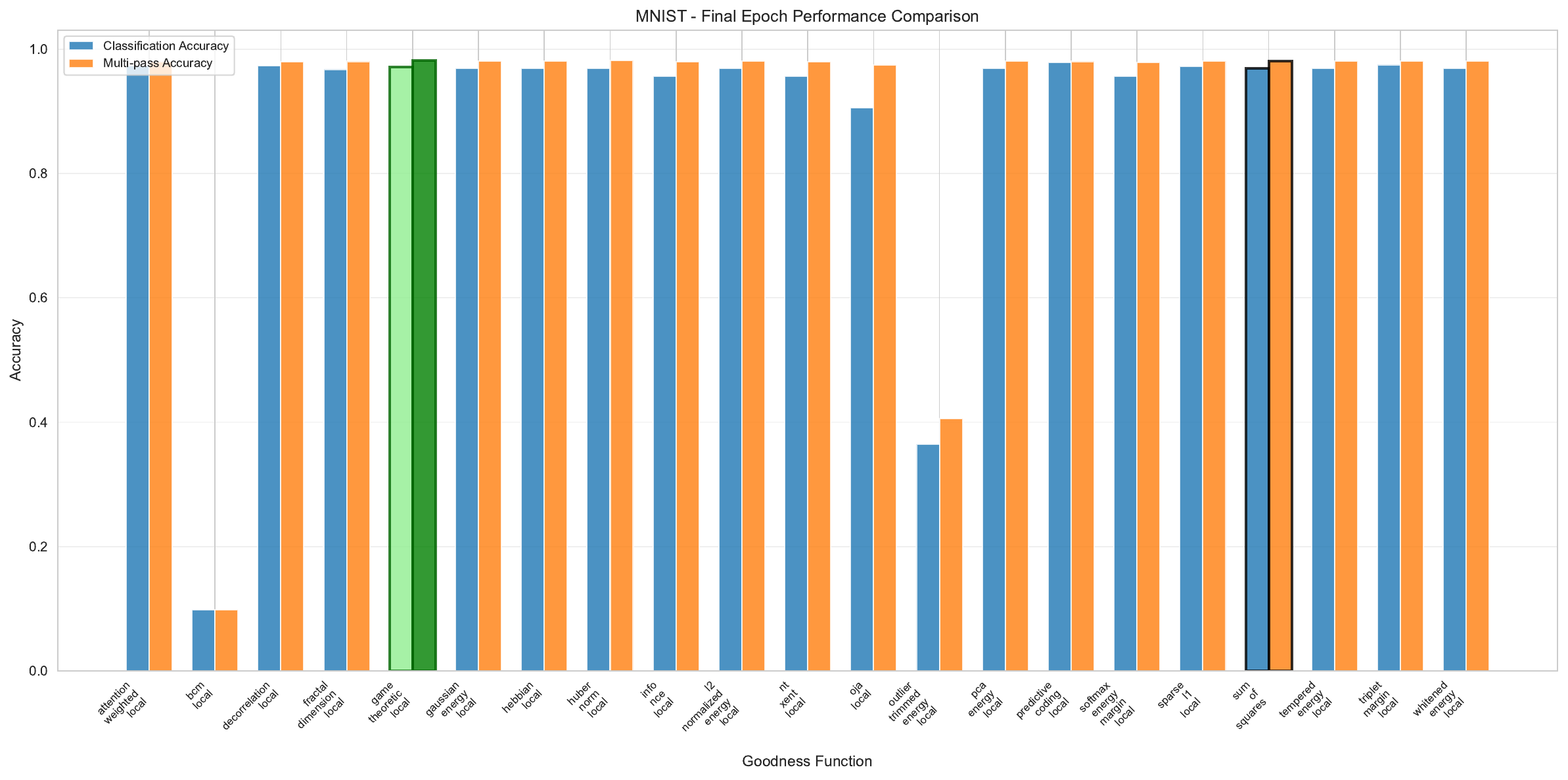}
    \caption{Comparison of final classification accuracy on MNIST. The majority of goodness functions achieve high accuracy ($>97\%$), with \texttt{game\_theoretic\_local} performing best. This indicates the Forward-Forward algorithm is highly effective for digit recognition across diverse objectives.}
    \label{fig:mnist_bars}
    \Description{A bar chart comparing Classification Accuracy and Multi-pass Accuracy on MNIST. Almost all functions show ceiling performance with bars reaching near 98 percent. Exceptions are BCM local and outlier trimmed energy local, which show significantly lower performance.}
\end{figure}

\subsection{Performance on STL-10}
STL-10 presents a challenge with its higher resolution and fewer labeled examples (Table \ref{tab:stl10_results}).

\textbf{Accuracy:} The \texttt{triplet\_margin\_local} function achieved the highest \textbf{Multi-pass Accuracy of 37.72\%}, outperforming the baseline (36.64\%). This suggests that for data with limited labels, explicitly enforcing separation between positive and negative samples via a margin is crucial for learning robust decision boundaries.

\textbf{Efficiency:} The \texttt{bcm\_local} function was the most energy-efficient (0.0112 kWh) but failed to learn useful representations (14.47\% multi-pass accuracy). Among the high-performing models, \texttt{triplet\_margin\_local} offered a balanced profile with 6.46 g CO$_2$ emissions.

\begin{table}[!htbp]
\centering
\caption{Performance metrics for the STL-10 dataset. \textbf{Class. Acc.} denotes the accuracy of a downstream linear classifier. \textbf{Multi-pass Acc.} refers to the native Forward-Forward inference accuracy. \textbf{Emissions} and \textbf{Energy} represent the total environmental cost of training. The best performing method for each metric is highlighted in bold.}
\label{tab:stl10_results}
\setlength{\tabcolsep}{5pt}
\renewcommand{\arraystretch}{1.3}
\begin{tabularx}{\linewidth}{@{}
  >{\RaggedRight\arraybackslash}p{4.5cm}%
  >{\centering\arraybackslash}Y%
  >{\centering\arraybackslash}Y%
  >{\centering\arraybackslash}Y%
  >{\centering\arraybackslash}Y%
  >{\centering\arraybackslash}Y@{}}
\toprule
\textbf{Goodness Function} & \textbf{Class. Acc.} & \textbf{Multi-pass Acc.} & \textbf{Class. Loss} & \textbf{Emissions} (g CO$_2$) & \textbf{Energy} (kWh) \\
\midrule
attention\_weighted\_local & 0.3647 & 0.3647 & 1.8791 & 6.4523 & 0.011740 \\
bcm\_local & 0.2132 & 0.1447 & 2.0937 & 6.1793 & 0.011244 \\
decorrelation\_local & 0.3649 & 0.3689 & 1.8583 & 6.4637 & 0.011761 \\
fractal\_dimension\_local & 0.3554 & 0.3614 & 1.8717 & 6.3251 & 0.011509 \\
game\_theoretic\_local & 0.3770 & 0.3561 & 1.8375 & 6.7173 & 0.012223 \\
\addlinespace[0.1cm]
gaussian\_energy\_local & 0.3655 & 0.3664 & 1.8829 & 6.7519 & 0.012285 \\
hebbian\_local & 0.3655 & 0.3664 & 1.8829 & 6.5574 & 0.011932 \\
huber\_norm\_local & 0.3699 & 0.3479 & 1.8647 & 6.6847 & 0.012163 \\
info\_nce\_local & 0.3494 & 0.3354 & 1.9545 & 6.6129 & 0.012033 \\
l2\_normalized\_energy\_local & 0.3655 & 0.3664 & 1.8829 & 6.4440 & 0.011725 \\
\addlinespace[0.1cm]
nt\_xent\_local & 0.3494 & 0.3354 & 1.9545 & 6.3271 & 0.011513 \\
oja\_local & 0.3632 & 0.1096 & 1.8595 & 6.4125 & 0.011668 \\
outlier\_trimmed\_energy\_local & 0.2764 & 0.1004 & 1.9473 & 6.7668 & 0.012313 \\
pca\_energy\_local & 0.3655 & 0.3664 & 1.8829 & 6.5554 & 0.011928 \\
predictive\_coding\_local & 0.4152 & 0.3607 & 1.6690 & 6.7336 & 0.012252 \\
\addlinespace[0.1cm]
softmax\_energy\_margin\_local & 0.3475 & 0.3231 & 1.8942 & 6.6606 & 0.012119 \\
sparse\_l1\_local & 0.3581 & 0.3657 & 1.8809 & 6.4395 & 0.011717 \\
\rowcolor{gray!15} \textit{sum\_of\_squares} (baseline) & 0.3655 & 0.3664 & 1.8829 & 10.1645 & 0.018495 \\
tempered\_energy\_local & 0.3655 & 0.3664 & 1.8829 & 6.2232 & 0.011323 \\
\rowcolor{green!10} \textbf{triplet\_margin\_local} & \textbf{0.3769} & \textbf{0.3772} & \textbf{1.8449} & \textbf{6.4643} & \textbf{0.011762} \\
\addlinespace[0.1cm]
whitened\_energy\_local & 0.3655 & 0.3664 & 1.8829 & 6.5533 & 0.011924 \\
\bottomrule
\end{tabularx}
\end{table}

\begin{figure}[!htbp]
    \centering
    \includegraphics[width=0.9\linewidth]{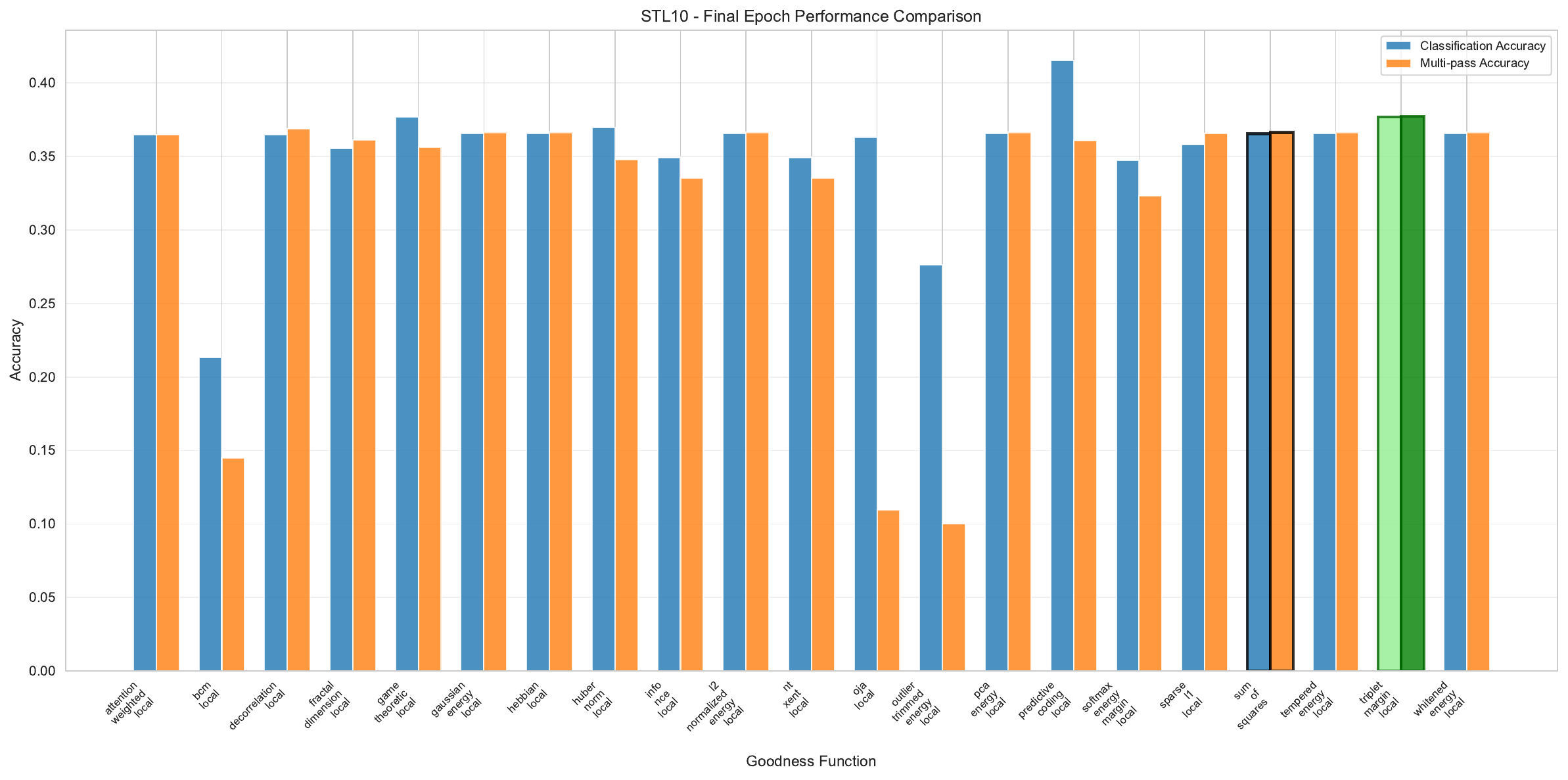}
    \caption{Comparison of final classification accuracy on STL-10. \texttt{triplet\_margin\_local} achieved the highest Multi-pass Accuracy (37.72\%), suggesting that explicit separation between positive and negative samples is crucial for data-scarce, high-resolution tasks.}
    \label{fig:stl10_bars}
    \Description{A bar chart comparing accuracy on STL-10. Overall accuracy is lower than other datasets, ranging between 30 and 40 percent for top performers. Triplet margin local has the tallest orange bar representing multi-pass accuracy.}
\end{figure}

Additional detailed figures including classification loss curves, layer-wise accuracy progression, and multi-pass accuracy analysis for all datasets are provided in the Appendix. For classification loss curves please refer to Figures \ref{fig:mnistloss}, \ref{fig:fashionloss}, \ref{fig:cifarloss}, \ref{fig:stlloss} in Appendix \ref{app:a}.For layer-wise accuracy progression please refer to Figures \ref{fig:mnist_perlayer}, \ref{fig:fashion_perlayer}, \ref{fig:cifar_perlayer},  \ref{fig:stl10_perlayer} in Appendix \ref{app:b}. For multi-pass accuracy analysis for all datasets, please refer to Figures \ref{fig:mnsit_multi}, \ref{fig:fashion_multi}, \ref{fig:cifar_multi}, \ref{fig:stl10_multi} in Appendix \ref{app:c}.
\section{Discussion}

In this work, we set out to explore the landscape of "goodness" functions for the Forward-Forward algorithm. Our benchmarking of 21 objectives across four datasets allows us to answer the research questions posed at the beginning of this study.

\subsection{RQ1: Impact on Classification Accuracy}
Our first question asked how the choice of goodness function impacts accuracy compared to the standard sum-of-squares baseline. The results unequivocally demonstrate that \textbf{minimizing prediction error and enforcing margins are superior to simple activity maximization.}

The \texttt{predictive\_coding\_local} function consistently outperformed (on classification accuracy) the baseline on almost all datasets including complex datasets like CIFAR-10, achieving a Multi-pass Accuracy of 43.42\% compared to the baseline's 39.59\%. This suggests that the standard Forward-Forward objective is insufficient for capturing the rich structure of natural images. By contrast, predictive coding forces the network to learn dependencies between neurons, aligning with theories of cortical function where the brain constantly attempts to predict its own states \cite{friston2010free}.

However, prediction is not the only path to goodness. On FashionMNIST, the \texttt{softmax\_energy\_margin\_local} function achieved the highest Multi-pass Accuracy (86.32\%), and on STL-10, \texttt{triplet\_margin\_local} led the pack (37.72\%). This indicates that for fine-grained or data-scarce tasks, objectives that explicitly enforce separation between positive and negative samples, similar to contrastive learning margins which are highly effective in the Forward-Forward regime.

\subsection{RQ2: Energy Consumption and Sustainability}
Our second question investigated the variability in energy consumption and sought more sustainable alternatives. We found that the "carbon cost" of learning is highly sensitive to the specific mathematical operations used in the goodness function.

We identified \texttt{whitened\_energy\_local} and \texttt{sparse\_l1\_local} as the most sustainable alternatives. On FashionMNIST, the whitened energy objective reduced carbon emissions by approximately 10\% compared to the baseline (11.91 g vs 13.12 g) while maintaining competitive accuracy. This efficiency likely stems from the decorrelation of neural activities, which prevents redundant firing and allows the network to converge to a solution with fewer total synaptic updates. This finding supports the hypothesis that metabolically efficient learning rules such as those that favor sparse and decorrelated codes, and are naturally aligned with Green AI principles \cite{laughlin2003communication}.

\subsection{RQ3: Trade-offs and Pareto Optimality}
Our final question asked if there is a trade-off between predictive performance and environmental efficiency. To answer this, we look for \textbf{Pareto-optimal} solutions, with choices where it is impossible to improve accuracy without increasing energy consumption, or vice versa.

Our analysis reveals distinct "sweet spots" on the Pareto frontier:
\begin{enumerate}
    \item \textbf{The Performance Specialist:} For applications where accuracy is paramount, \texttt{sparse\_l1\_local} (CIFAR-10) and \texttt{softmax\_energy\_margin\_local} (FashionMNIST) are the clear choices. Notably, \texttt{sparse\_l1\_local} dominates on CIFAR-10 by being both the most accurate and the most efficient.
    \item \textbf{The Efficiency Specialist:} For resource-constrained environments, \texttt{whitened\_energy\_local} offers the best balance on simpler datasets like FashionMNIST. However, on complex data (CIFAR-10), \texttt{sparse\_l1\_local} is the optimal choice for both performance and efficiency.
\end{enumerate}

Crucially, the standard \texttt{sum\_of\_squares} baseline is often \textbf{sub-optimal}. In many cases, it is dominated by other functions that are either more accurate (like predictive coding) or more efficient (like whitened energy). This indicates that the default objective used in early Forward-Forward research is likely not the most effective choice for either performance or sustainability.

\subsection{Limitations, Failure Modes and Conclusion}
While many goodness functions demonstrated strong performance, we must acknowledge that certain objectives failed to learn meaningful representations. Specifically, functions like \texttt{bcm\_local} and \texttt{outlier\_trimmed\_energy\_local} occasionally reported accuracies around 10\% (e.g., 9.86\% for BCM on MNIST). Since these datasets contain 10 balanced classes, a 10\% accuracy corresponds to random chance, indicating a complete failure of the learning process.

We hypothesize that these failures stem from the inherent instability of these objectives in a greedy, layer-wise training regime. The BCM rule, for instance, relies on a sliding threshold that is sensitive to the statistics of the input stream. Without the global error signal of backpropagation to correct for "runaway" updates, the threshold may fail to stabilize, leading to representational collapse where all neurons either saturate or die. Similarly, outlier trimming can be too aggressive, discarding useful signal as noise. These results highlight a critical limitation of biologically plausible local learning: it requires careful hyperparameter tuning and homeostatic mechanisms to maintain stability, which global optimization methods often take for granted.

To conclude, the "search for goodness" reveals that the objective function is a powerful lever for shaping both the intelligence and the efficiency of biologically plausible networks. By moving beyond simple activity maximization and embracing principles of prediction, sparsity, and margins, we can build Forward-Forward networks that are not only smarter but also greener. Future work should extend this analysis to neuromorphic hardware, where the local nature of these rules could yield even greater efficiency gains \cite{davies2018loihi}.


\section*{Acknowledgments}
We acknowledge National Supercomputing 
Mission (NSM) for providing computing resources of ‘PARAM Ananta’ at IIT Gandhinagar, 
which is implemented by C-DAC and supported by the Ministry of Electronics and 
Information Technology (MeitY) and Department of Science and Technology (DST), 
Government of India.

\bibliographystyle{unsrt}  
\bibliography{references}  

\begin{thebibliography}{10}

\bibitem{rumelhart1986learning}
David~E Rumelhart, Geoffrey~E Hinton, and Ronald~J Williams.
\newblock Learning representations by back-propagating errors.
\newblock {\em Nature}, 323(6088):533--536, October 1986.

\bibitem{crick1989recent}
F~Crick.
\newblock The recent excitement about neural networks.
\newblock {\em Nature}, 337(6203):129--132, January 1989.

\bibitem{bengio2015towards}
Yoshua Bengio, Dong-Hyun Lee, Jorg Bornschein, Thomas Mesnard, and Zhouhan Lin.
\newblock Towards biologically plausible deep learning, 2016.

\bibitem{lillicrap2020backpropagation}
Timothy~P Lillicrap, Adam Santoro, Luke Marris, Colin~J Akerman, and Geoffrey Hinton.
\newblock Backpropagation and the brain.
\newblock {\em Nat. Rev. Neurosci.}, 21(6):335--346, June 2020.

\bibitem{whittington2019theories}
James C~R Whittington and Rafal Bogacz.
\newblock Theories of error back-propagation in the brain.
\newblock {\em Trends Cogn. Sci.}, 23(3):235--250, March 2019.

\bibitem{lillicrap2016random}
Timothy~P Lillicrap, Daniel Cownden, Douglas~B Tweed, and Colin~J Akerman.
\newblock Random synaptic feedback weights support error backpropagation for deep learning.
\newblock {\em Nat. Commun.}, 7(1):13276, November 2016.

\bibitem{scellier2017equilibrium}
Benjamin Scellier and Yoshua Bengio.
\newblock Equilibrium propagation: Bridging the gap between energy-based models and backpropagation.
\newblock {\em Front. Comput. Neurosci.}, 11:24, May 2017.

\bibitem{rao1999predictive}
R~P Rao and D~H Ballard.
\newblock Predictive coding in the visual cortex: a functional interpretation of some extra-classical receptive-field effects.
\newblock {\em Nat. Neurosci.}, 2(1):79--87, January 1999.

\bibitem{hinton2022forward}
Geoffrey Hinton.
\newblock The forward-forward algorithm: Some preliminary investigations, 2022.

\bibitem{chen2020simple}
Ting Chen, Simon Kornblith, Mohammad Norouzi, and Geoffrey Hinton.
\newblock A simple framework for contrastive learning of visual representations, 2020.

\bibitem{he2020momentum}
Kaiming He, Haoqi Fan, Yuxin Wu, Saining Xie, and Ross Girshick.
\newblock Momentum contrast for unsupervised visual representation learning, 2020.

\bibitem{oord2018representation}
Aaron van~den Oord, Yazhe Li, and Oriol Vinyals.
\newblock Representation learning with contrastive predictive coding, 2019.

\bibitem{strubell2019energy}
Emma Strubell, Ananya Ganesh, and Andrew McCallum.
\newblock Energy and policy considerations for deep learning in nlp, 2019.

\bibitem{schwartz2020green}
Roy Schwartz, Jesse Dodge, Noah~A Smith, and Oren Etzioni.
\newblock Green {AI}.
\newblock {\em Commun. ACM}, 63(12):54--63, November 2020.

\bibitem{patterson2021carbon}
David Patterson, Joseph Gonzalez, Quoc Le, Chen Liang, Lluis-Miquel Munguia, Daniel Rothchild, David So, Maud Texier, and Jeff Dean.
\newblock Carbon emissions and large neural network training, 2021.

\bibitem{hebb1949organization}
D.~O. Hebb.
\newblock {\em The organization of behavior; a neuropsychological theory.}
\newblock The organization of behavior; a neuropsychological theory. Wiley, Oxford, England, 1949.

\bibitem{oja1982simplified}
Erkki Oja.
\newblock Simplified neuron model as a principal component analyzer.
\newblock {\em J. Math. Biol.}, 15(3):267--273, November 1982.

\bibitem{bienenstock1982theory}
E~L Bienenstock, L~N Cooper, and P~W Munro.
\newblock Theory for the development of neuron selectivity: orientation specificity and binocular interaction in visual cortex.
\newblock {\em J. Neurosci.}, 2(1):32--48, January 1982.

\bibitem{bengio2014auto}
Yoshua Bengio.
\newblock How auto-encoders could provide credit assignment in deep networks via target propagation, 2014.

\bibitem{lee2015difference}
Dong-Hyun Lee, Saizheng Zhang, Asja Fischer, and Yoshua Bengio.
\newblock Difference target propagation, 2015.

\bibitem{jaderberg2017decoupled}
Max Jaderberg, Wojciech~Marian Czarnecki, Simon Osindero, Oriol Vinyals, Alex Graves, David Silver, and Koray Kavukcuoglu.
\newblock Decoupled neural interfaces using synthetic gradients, 2017.

\bibitem{lecun2006tutorial}
Yann LeCun, Sumit Chopra, Raia Hadsell, Aurelio Ranzato, and Fu~Jie Huang.
\newblock A tutorial on energy-based learning, 2006.

\bibitem{hopfield1982neural}
J~J Hopfield.
\newblock Neural networks and physical systems with emergent collective computational abilities.
\newblock {\em Proc. Natl. Acad. Sci. U. S. A.}, 79(8):2554--2558, April 1982.

\bibitem{hinton1983optimal}
Geoffrey~E. Hinton and Terrence~J. Sejnowski.
\newblock Optimal perceptual inference.
\newblock In {\em Proceedings of the {IEEE} Conference on Computer Vision and Pattern Recognition}, 1983.

\bibitem{hinton2002training}
Geoffrey~E Hinton.
\newblock Training products of experts by minimizing contrastive divergence.
\newblock {\em Neural Comput.}, 14(8):1771--1800, August 2002.

\bibitem{hyvarinen2005estimation}
Aapo Hyv{{\"a}}rinen.
\newblock Estimation of non-normalized statistical models by score matching.
\newblock {\em Journal of Machine Learning Research}, 6(24):695--709, 2005.

\bibitem{grill2020bootstrap}
Jean-Bastien Grill, Florian Strub, Florent Altch\'{e}, Corentin Tallec, Pierre Richemond, Elena Buchatskaya, Carl Doersch, Bernardo Avila~Pires, Zhaohan Guo, Mohammad Gheshlaghi~Azar, Bilal Piot, koray kavukcuoglu, Remi Munos, and Michal Valko.
\newblock Bootstrap your own latent - a new approach to self-supervised learning.
\newblock In H.~Larochelle, M.~Ranzato, R.~Hadsell, M.F. Balcan, and H.~Lin, editors, {\em Advances in Neural Information Processing Systems}, volume~33, pages 21271--21284. Curran Associates, Inc., 2020.

\bibitem{chen2021exploring}
Xinlei Chen and Kaiming He.
\newblock Exploring simple siamese representation learning, 2020.

\bibitem{zbontar2021barlow}
Jure Zbontar, Li~Jing, Ishan Misra, Yann LeCun, and Stéphane Deny.
\newblock Barlow twins: Self-supervised learning via redundancy reduction, 2021.

\bibitem{caron2020unsupervised}
Mathilde Caron, Ishan Misra, Julien Mairal, Priya Goyal, Piotr Bojanowski, and Armand Joulin.
\newblock Unsupervised learning of visual features by contrasting cluster assignments, 2021.

\bibitem{schroff2015facenet}
Florian Schroff, Dmitry Kalenichenko, and James Philbin.
\newblock Facenet: A unified embedding for face recognition and clustering.
\newblock In {\em 2015 IEEE Conference on Computer Vision and Pattern Recognition (CVPR)}, page 815–823. IEEE, June 2015.

\bibitem{cogswell2015reducing}
Michael Cogswell, Faruk Ahmed, Ross Girshick, Larry Zitnick, and Dhruv Batra.
\newblock Reducing overfitting in deep networks by decorrelating representations, 2016.

\bibitem{balduzzi2018mechanics}
David Balduzzi, Sebastien Racaniere, James Martens, Jakob Foerster, Karl Tuyls, and Thore Graepel.
\newblock The mechanics of n-player differentiable games, 2018.

\bibitem{mandelbrot1982fractal}
M~J Kirkby.
\newblock The fractal geometry of nature. benoit b. mandelbrot. w. h. freeman and co., san francisco, 1982. no. of pages: 460. price: \pounds{}22.75 (hardback).
\newblock {\em Earth Surf. Process.}, 8(4):406--406, July 1983.

\bibitem{huber1964robust}
Peter~J Huber.
\newblock Robust estimation of a location parameter.
\newblock In {\em Springer Series in Statistics}, Springer series in statistics, pages 492--518. Springer New York, New York, NY, 1992.

\bibitem{rousseeuw1984least}
Peter~J Rousseeuw.
\newblock Least median of squares regression.
\newblock {\em J. Am. Stat. Assoc.}, 79(388):871, December 1984.

\bibitem{liu2016large}
Weiyang Liu, Yandong Wen, Zhiding Yu, and Meng Yang.
\newblock Large-margin softmax loss for convolutional neural networks, 2017.

\bibitem{hyvarinen2000independent}
A~Hyv{\"a}rinen and E~Oja.
\newblock Independent component analysis: algorithms and applications.
\newblock {\em Neural Netw.}, 13(4-5):411--430, May 2000.

\bibitem{olshausen1996emergence}
B~A Olshausen and D~J Field.
\newblock Emergence of simple-cell receptive field properties by learning a sparse code for natural images.
\newblock {\em Nature}, 381(6583):607--609, June 1996.

\bibitem{vaswani2017attention}
Ashish Vaswani, Noam Shazeer, Niki Parmar, Jakob Uszkoreit, Llion Jones, Aidan~N. Gomez, Lukasz Kaiser, and Illia Polosukhin.
\newblock Attention is all you need, 2023.

\bibitem{broomhead1988multivariable}
David Broomhead and David Lowe.
\newblock Radial basis functions, multi-variable functional interpolation and adaptive networks.
\newblock {\em ROYAL SIGNALS AND RADAR ESTABLISHMENT MALVERN (UNITED KINGDOM)}, RSRE-MEMO-4148, 03 1988.

\bibitem{codecarbon}
Beno{\^\i}t Courty, Victor Schmidt, {Goyal-Kamal}, {inimaz}, {MarionCoutarel}, Luis Blanche, Boris Feld, J{\'e}r{\'e}my Lecourt, {LiamConnell}, Amine Saboni, {SabAmine}, {supatomic}, Patrick Lloret, Mathilde L{\'e}val, Alexis Cruveiller, {ouminasara}, Franklin Zhao, Christian Bauer, Aditya Joshi, Jerry~Laruba Festus, Alexis Bogroff, Niko Laskaris, Hugues de~Lavoreille, Alexandre Phiev, Edoardo Abati, Douglas Blank, {rosekelly}, and {cianc}.
\newblock mlco2/codecarbon: v3.1.0, 2025.

\bibitem{friston2010free}
Karl Friston.
\newblock The free-energy principle: a unified brain theory?
\newblock {\em Nat. Rev. Neurosci.}, 11(2):127--138, February 2010.

\bibitem{laughlin2003communication}
Simon~B Laughlin and Terrence~J Sejnowski.
\newblock Communication in neuronal networks.
\newblock {\em Science}, 301(5641):1870--1874, September 2003.

\bibitem{davies2018loihi}
Mike Davies, Narayan Srinivasa, Tsung-Han Lin, Gautham Chinya, Yongqiang Cao, Sri~Harsha Choday, Georgios Dimou, Prasad Joshi, Nabil Imam, Shweta Jain, Yuyun Liao, Chit-Kwan Lin, Andrew Lines, Ruokun Liu, Deepak Mathaikutty, Steven McCoy, Arnab Paul, Jonathan Tse, Guruguhanathan Venkataramanan, Yi-Hsin Weng, Andreas Wild, Yoonseok Yang, and Hong Wang.
\newblock Loihi: A neuromorphic manycore processor with on-chip learning.
\newblock {\em IEEE Micro}, 38(1):82--99, 2018.

\end{thebibliography}

\appendix
\section{Appendix A. Per Dataset Classification Loss Plots}
\label{app:a}
\begin{figure}[!htbp]
    \centering
    \includegraphics[width=0.9\linewidth]{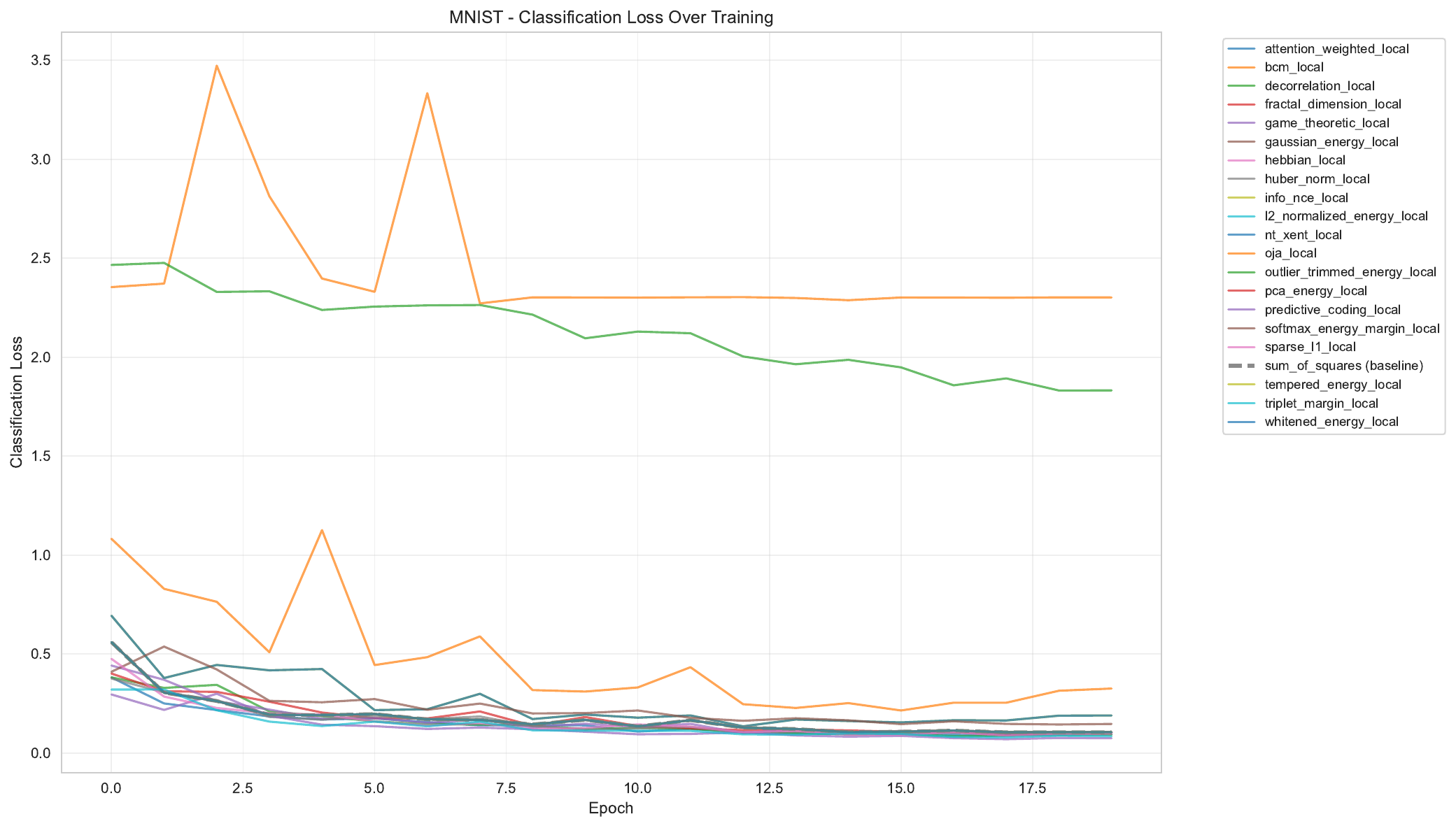}
    \caption{Classification Loss over training epochs on the MNIST dataset. Most functions converge rapidly to a low loss, while unstable functions like \texttt{bcm\_local} exhibit high, fluctuating loss values.}
    \label{fig:mnistloss}
    \Description{Line graph showing classification loss over 20 epochs for MNIST. Most lines flatline near zero very quickly. Two lines (representing BCM and outlier trimmed energy) remain high and erratic.}
\end{figure}
\begin{figure}[!htbp]
    \centering
    \includegraphics[width=0.9\linewidth]{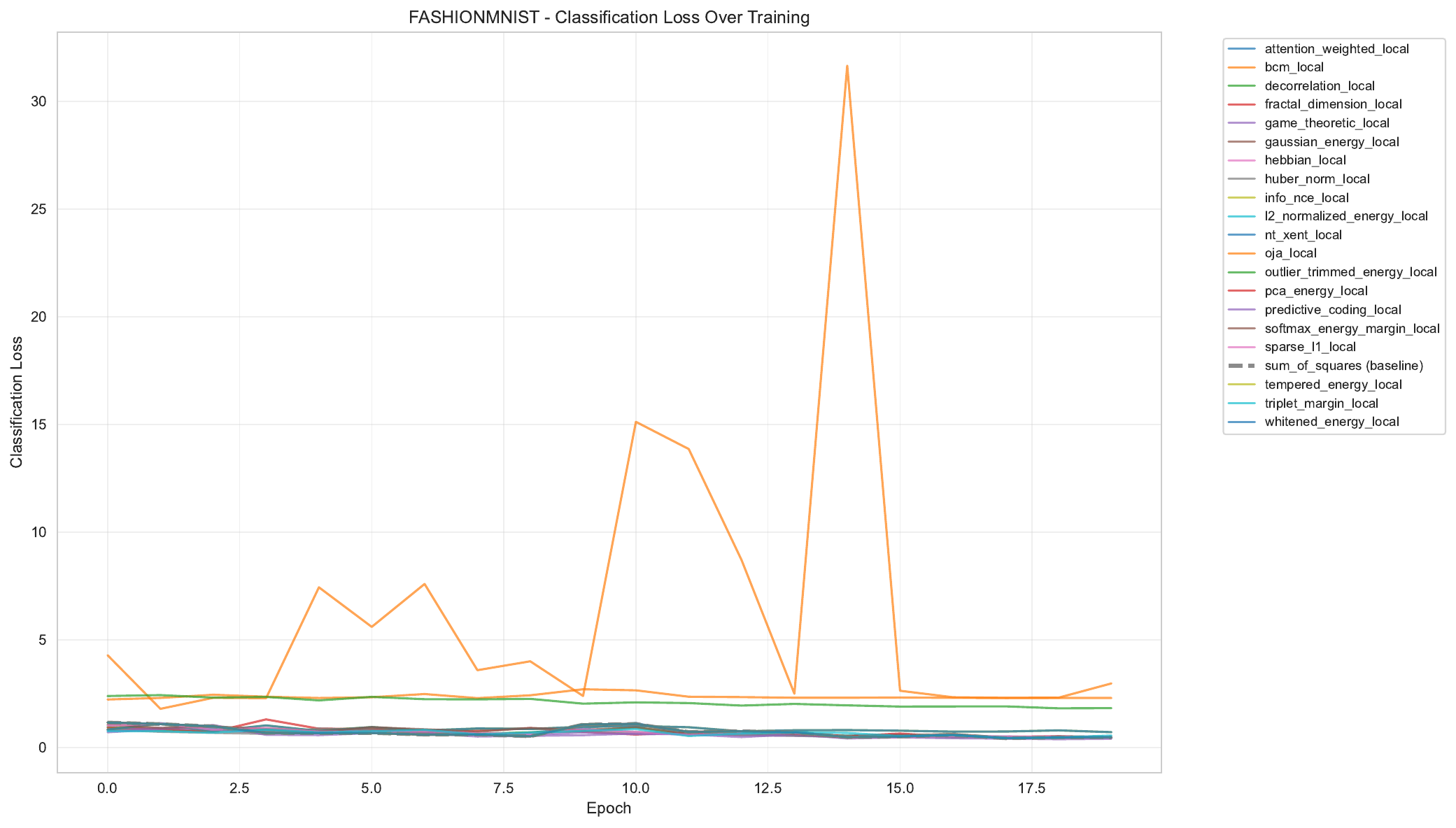}
    \caption{Classification Loss over training epochs on the FashionMNIST dataset. \texttt{triplet\_margin\_local} shows a distinct spike in loss during mid-training before converging.}
    \label{fig:fashionloss}
    \Description{Line graph showing classification loss over 20 epochs for FashionMNIST. Most lines are low and stable. One line shows a significant spike around epoch 14 before returning to lower values.}
\end{figure}
\begin{figure}[!htbp]
    \centering
    \includegraphics[width=0.9\linewidth]{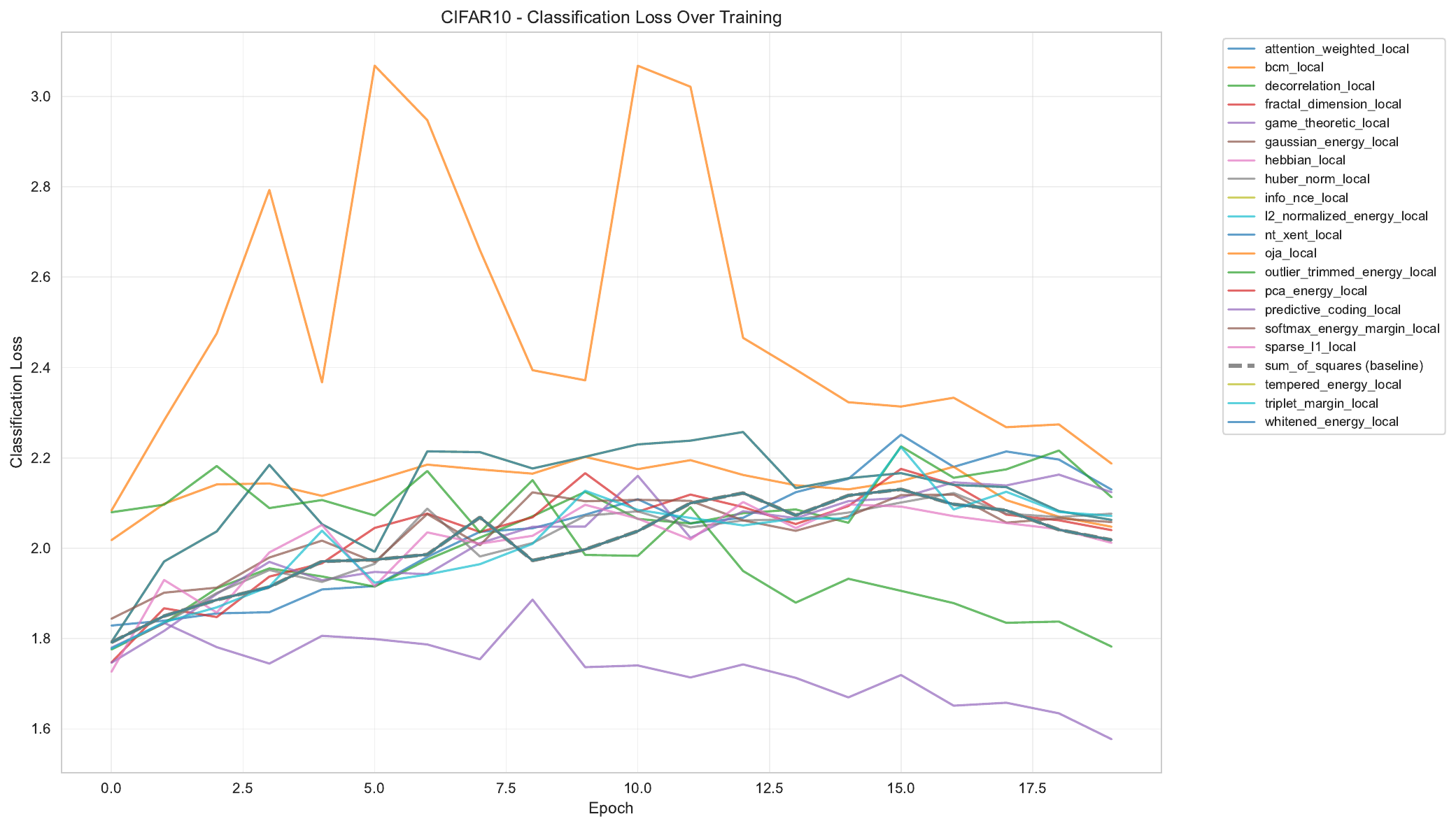}
    \caption{Classification Loss over training epochs on the CIFAR-10 dataset. The loss curves exhibit more variance compared to MNIST, reflecting the higher complexity of the dataset.}
    \label{fig:cifarloss}
    \Description{Line graph showing classification loss for CIFAR-10. The lines are more spread out and show more fluctuation compared to simple datasets, generally trending downwards but with visible noise.}
\end{figure}
\begin{figure}[!htbp]
    \centering
    \includegraphics[width=0.9\linewidth]{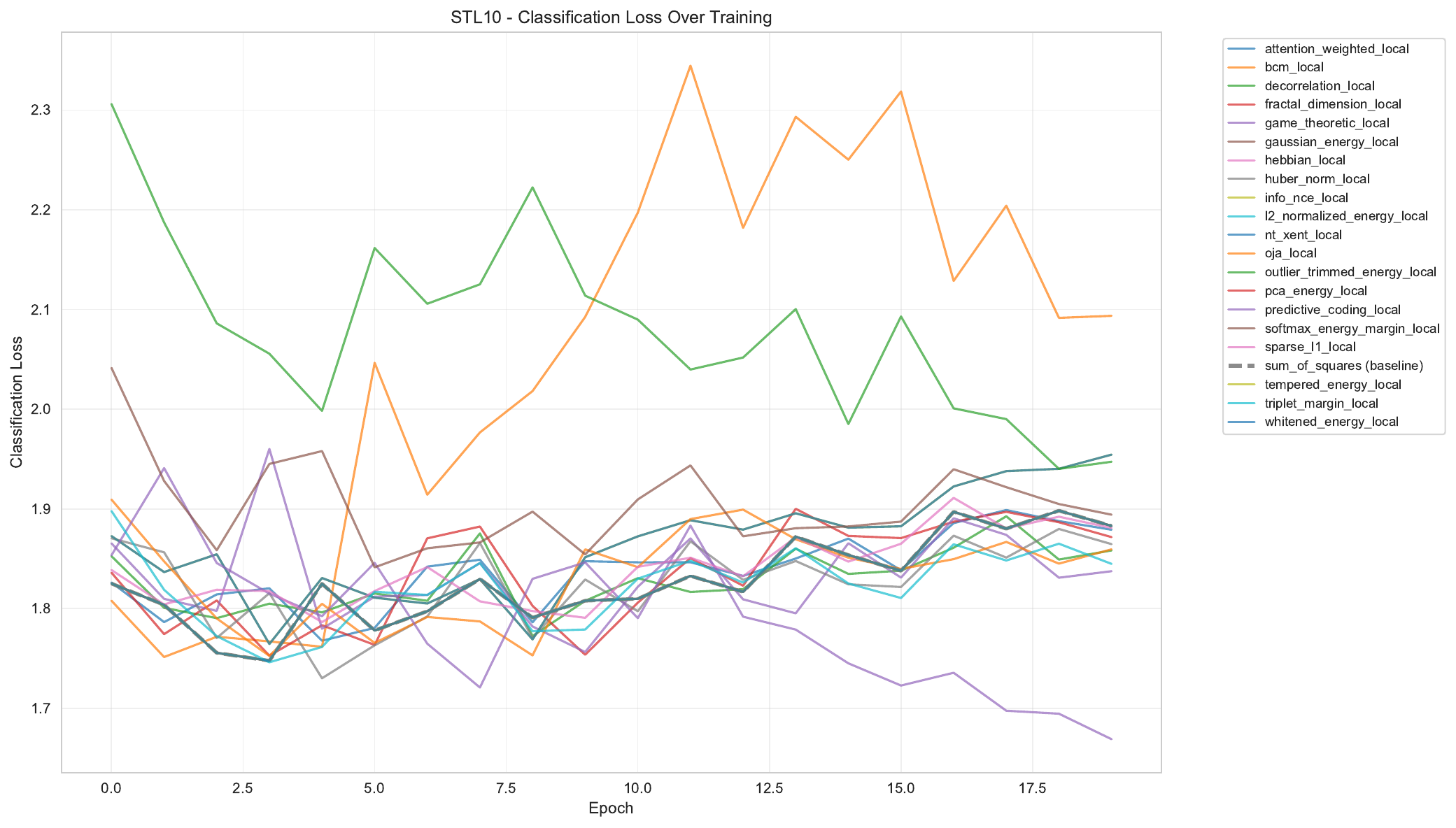} 
    \caption{Classification Loss over training epochs on the STL-10 dataset. Several functions struggle to minimize loss consistently, indicated by jagged and non-converging lines.}
    \label{fig:stlloss}
    \Description{Line graph showing classification loss for STL-10. The graph is noisy with several lines showing erratic spikes and lack of smooth convergence, indicating training difficulty.}
\end{figure}
\section{Appendix B. Per Dataset Accuracy Per Layer Plots}
\label{app:b}
\begin{figure*}[!htbp]
    \centering
    \includegraphics[width=\linewidth]{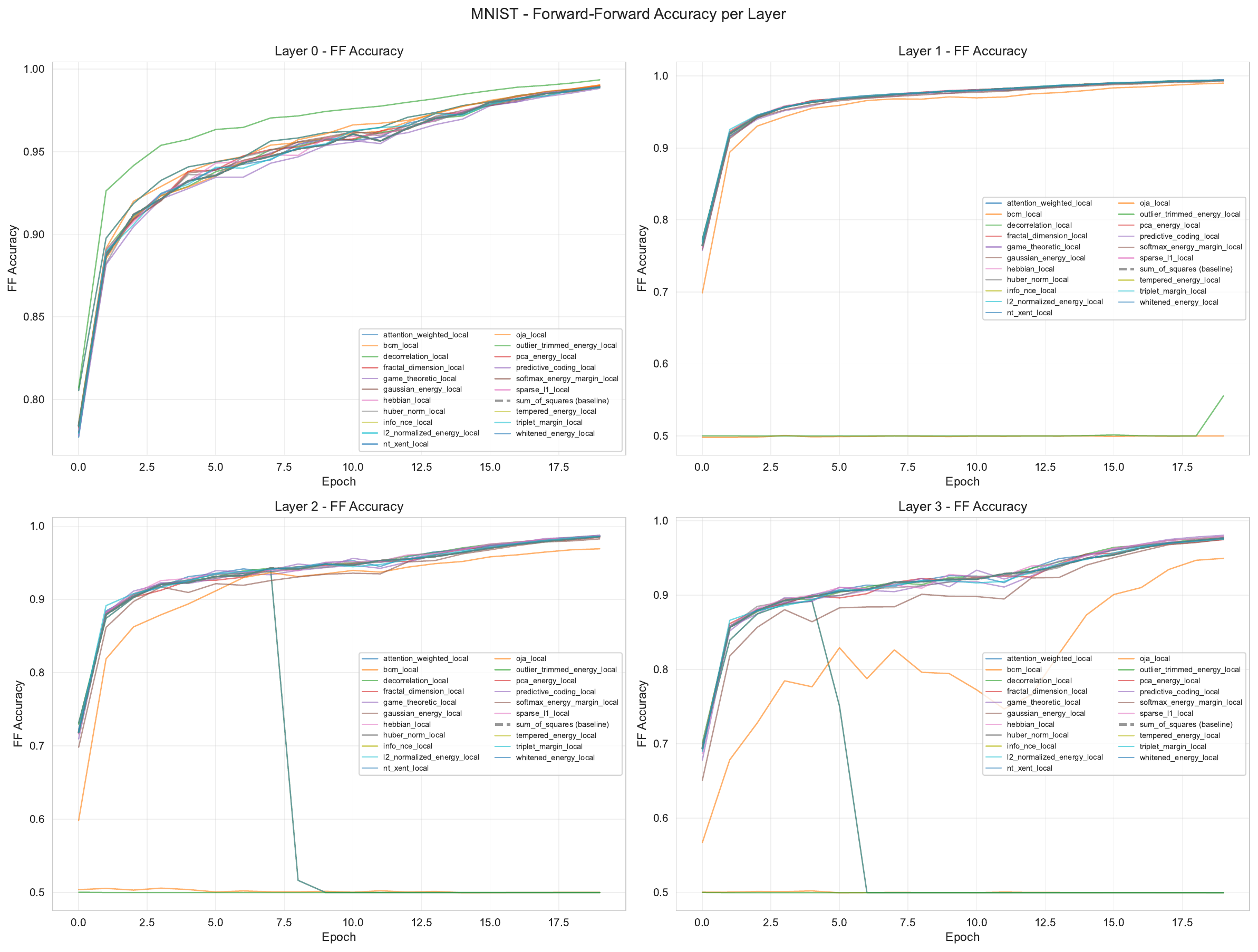} 
    \caption{Forward-Forward Accuracy per Layer on MNIST. Panels display the accuracy progression for Layer 0 through Layer 3. Deeper layers generally maintain high accuracy, though some functions show degradation in Layer 3.}
    \label{fig:mnist_perlayer}
    \Description{Four subplots showing accuracy curves for Layers 0, 1, 2, and 3 on MNIST. In all layers, most functions rapidly rise to near 1.0 accuracy. Layer 2 and 3 show one function dropping to zero.}
\end{figure*}
\begin{figure*}[!htbp]
    \centering
    \includegraphics[width=\linewidth]{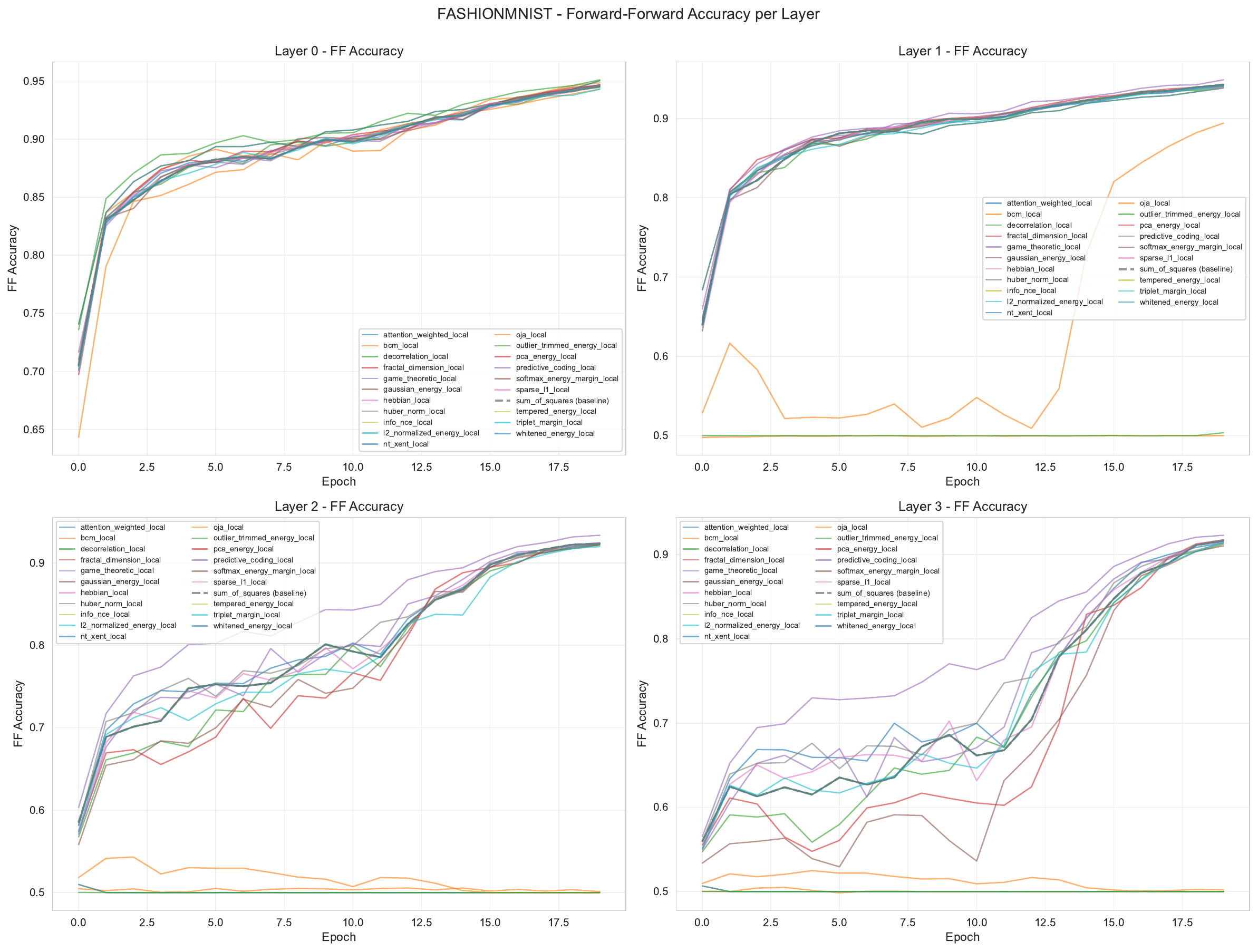}
    \caption{Forward-Forward Accuracy per Layer on FashionMNIST. Accuracy tends to increase in deeper layers for top-performing functions, while unstable functions fail to improve after the first layer.}
    \label{fig:fashion_perlayer}
    \Description{Four subplots showing accuracy curves for Layers 0, 1, 2, and 3 on FashionMNIST. Layer 0 shows a tight grouping of lines rising to 0.9. Deeper layers show more spread between the functions.}
\end{figure*}
\begin{figure*}[!htbp]
    \centering
    \includegraphics[width=\linewidth]{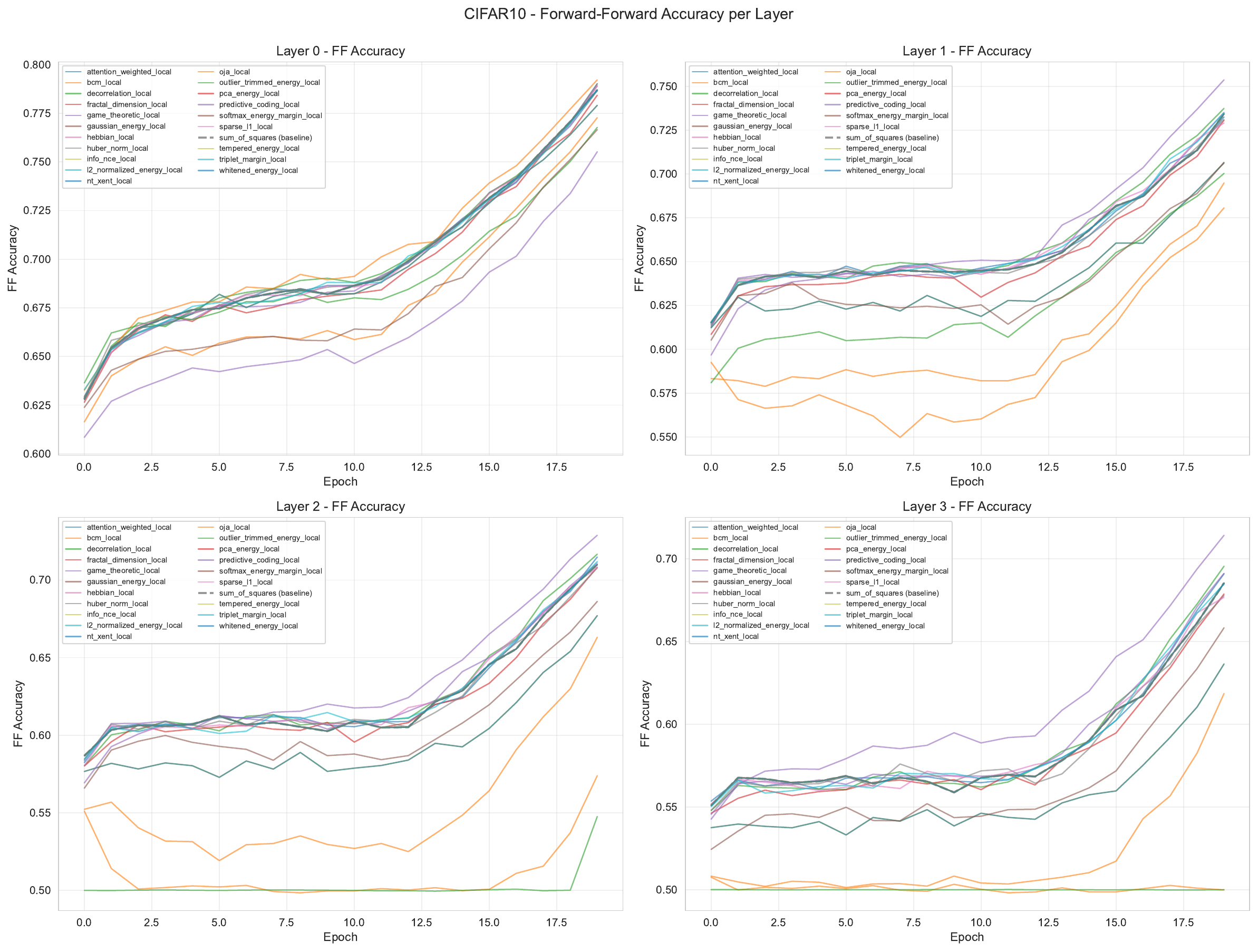}
    \caption{Forward-Forward Accuracy per Layer on CIFAR-10. The gap between high-performing functions (like predictive coding) and baselines becomes more pronounced in deeper network layers.}
    \label{fig:cifar_perlayer}
\end{figure*}
\begin{figure*}[!htbp]
    \centering
    \includegraphics[width=\linewidth]{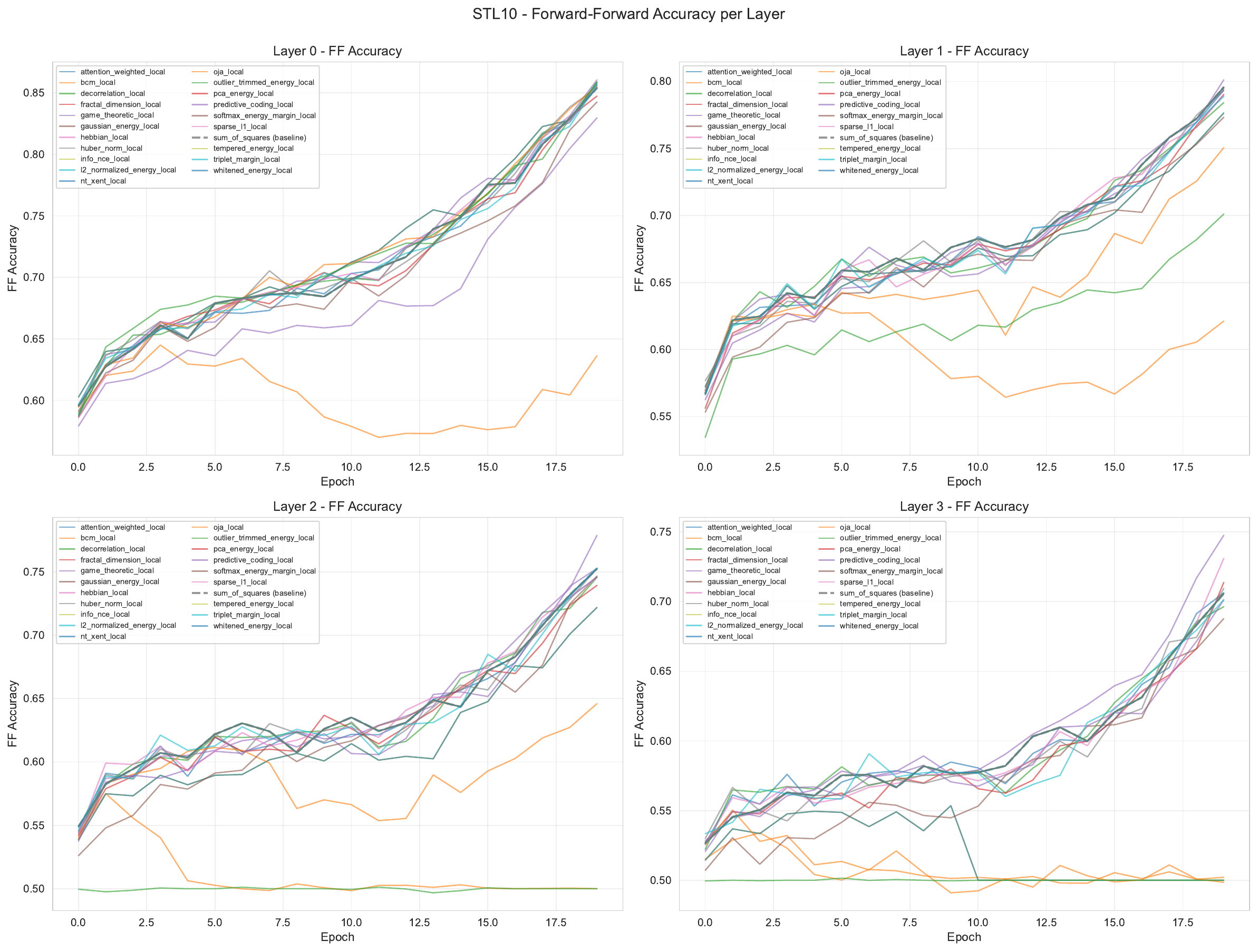}
    \caption{Forward-Forward Accuracy per Layer on STL-10. Accuracy progression is slower and more variable compared to simpler datasets, highlighting the difficulty of the task.}
    \label{fig:stl10_perlayer}
\end{figure*}
\section{Appendix C. Per Dataset Multi-pass Accuracy Plots}
\label{app:c}
\begin{figure}[!htbp]
    \centering
    \includegraphics[width=0.9\linewidth]{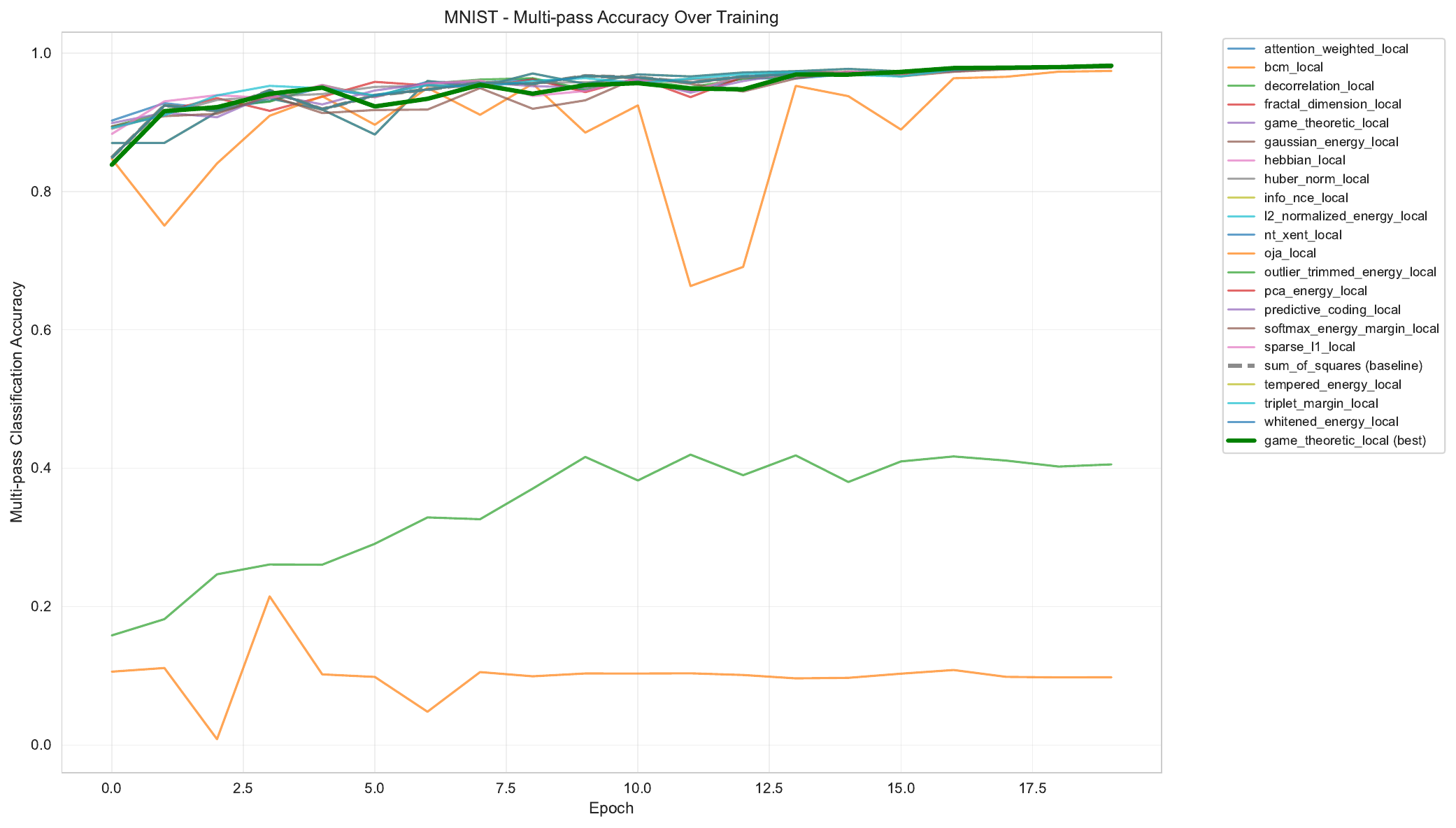}
    \caption{Multi-pass Accuracy progression over training epochs on MNIST. High stability is observed for \texttt{game\_theoretic\_local} (green line).}
    \label{fig:mnsit_multi}
    \Description{Line graph of multi-pass accuracy for MNIST. Most lines cluster tightly at the top of the graph (high accuracy). One line dips significantly in the middle epochs before recovering.}
\end{figure}
\begin{figure}[!htbp]
    \centering
    \includegraphics[width=0.9\linewidth]{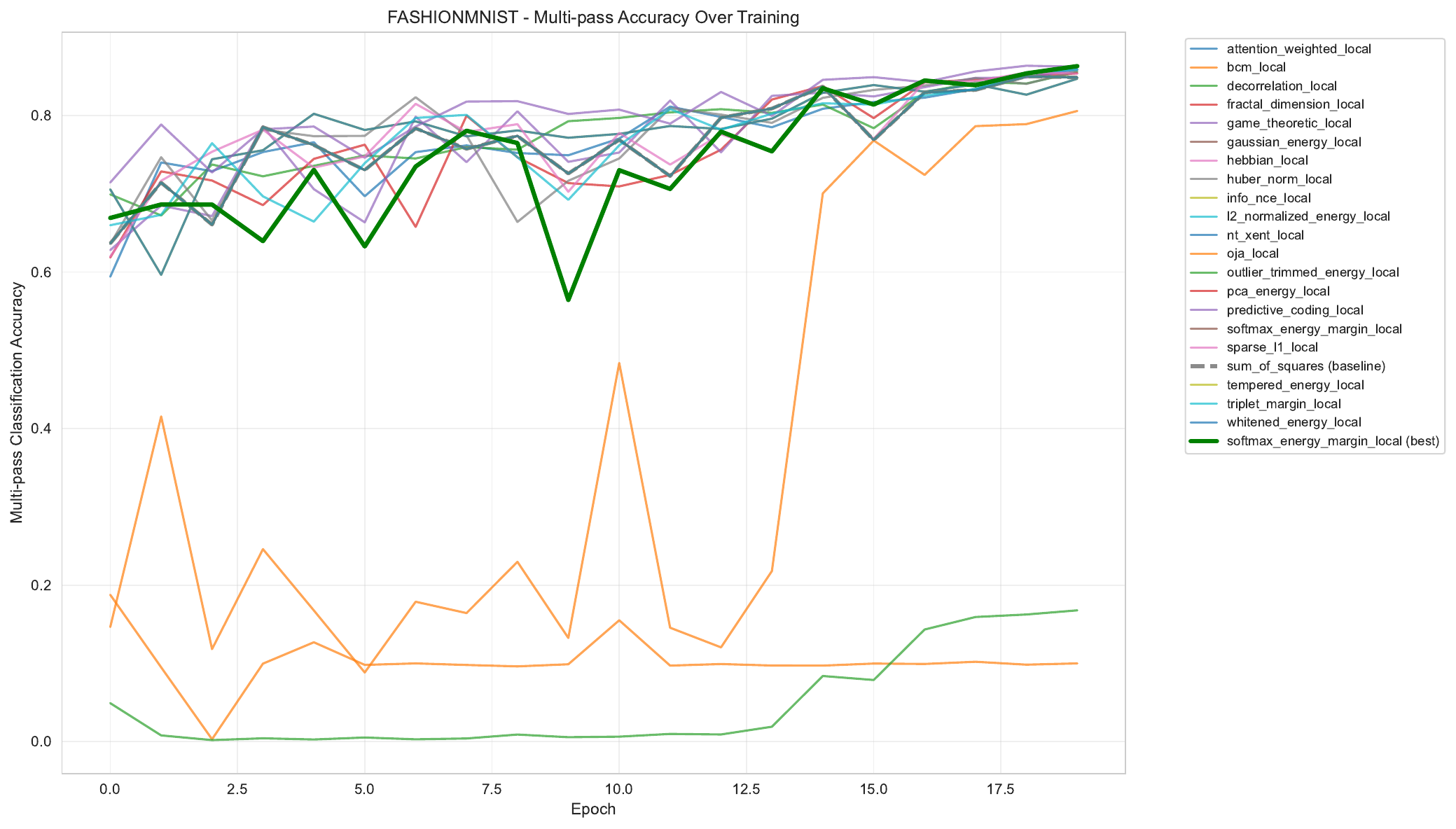}
    \caption{Multi-pass Accuracy progression over training epochs on FashionMNIST. The \texttt{softmax\_energy\_margin\_local} function (highlighted green) demonstrates consistent improvement and stability.}
    \label{fig:fashion_multi}
    \Description{Line graph of multi-pass accuracy for FashionMNIST. The green line representing softmax energy margin consistently trends upwards, ending near the top of the cluster.}
\end{figure}
\begin{figure}[!htbp]
    \centering
    \includegraphics[width=0.9\linewidth]{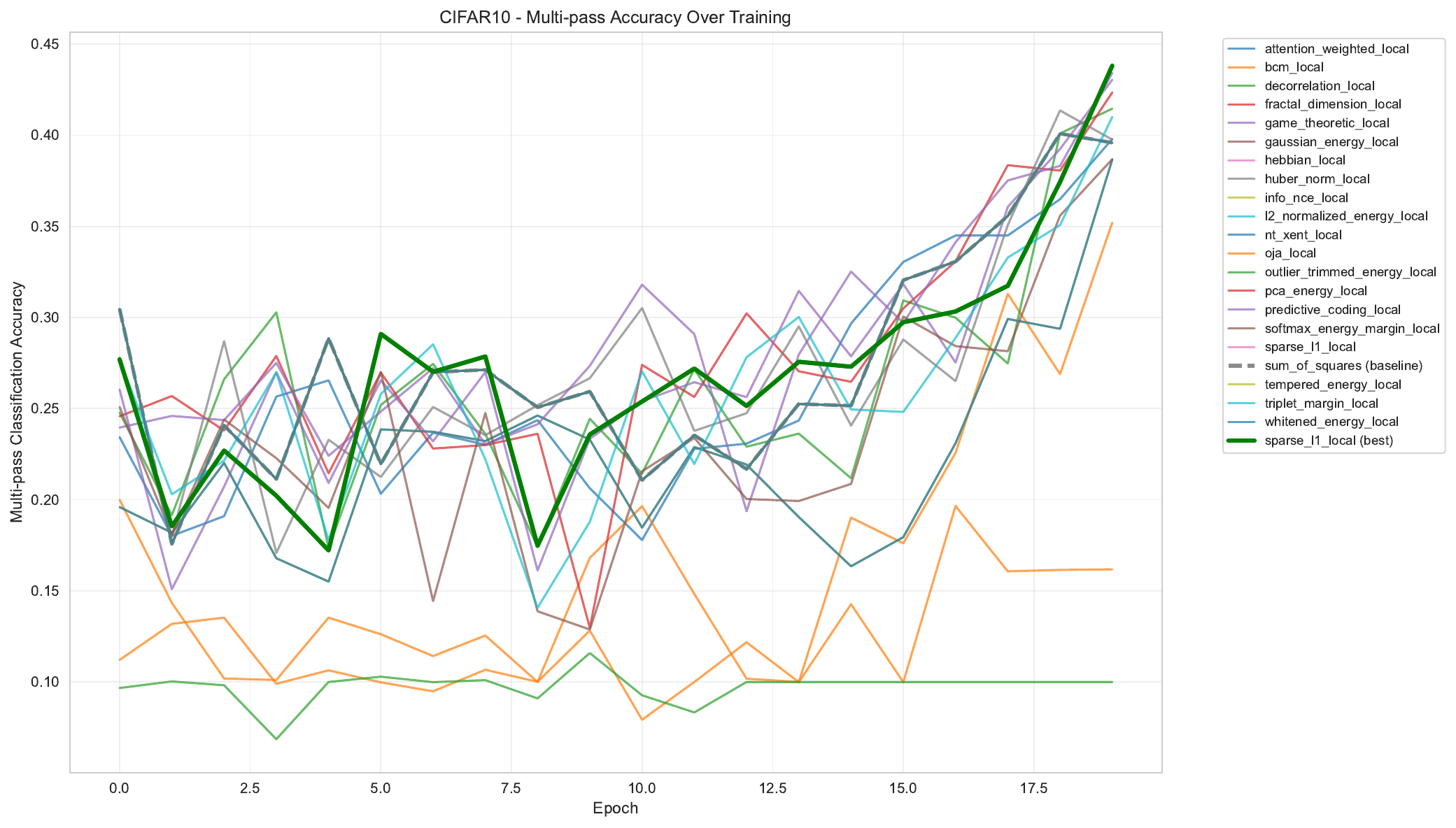}
    \caption{Multi-pass Accuracy progression over training epochs on CIFAR-10. \texttt{sparse\_l1\_local} (green line) shows a steady upward trend, outperforming the baseline.}
    \label{fig:cifar_multi}
    \Description{Line graph of multi-pass accuracy for CIFAR-10. The lines are volatile. The green line for sparse L1 shows a strong positive slope in later epochs, separating from the pack.}
\end{figure}
\begin{figure}[!htbp]
    \centering
    \includegraphics[width=0.9\linewidth]{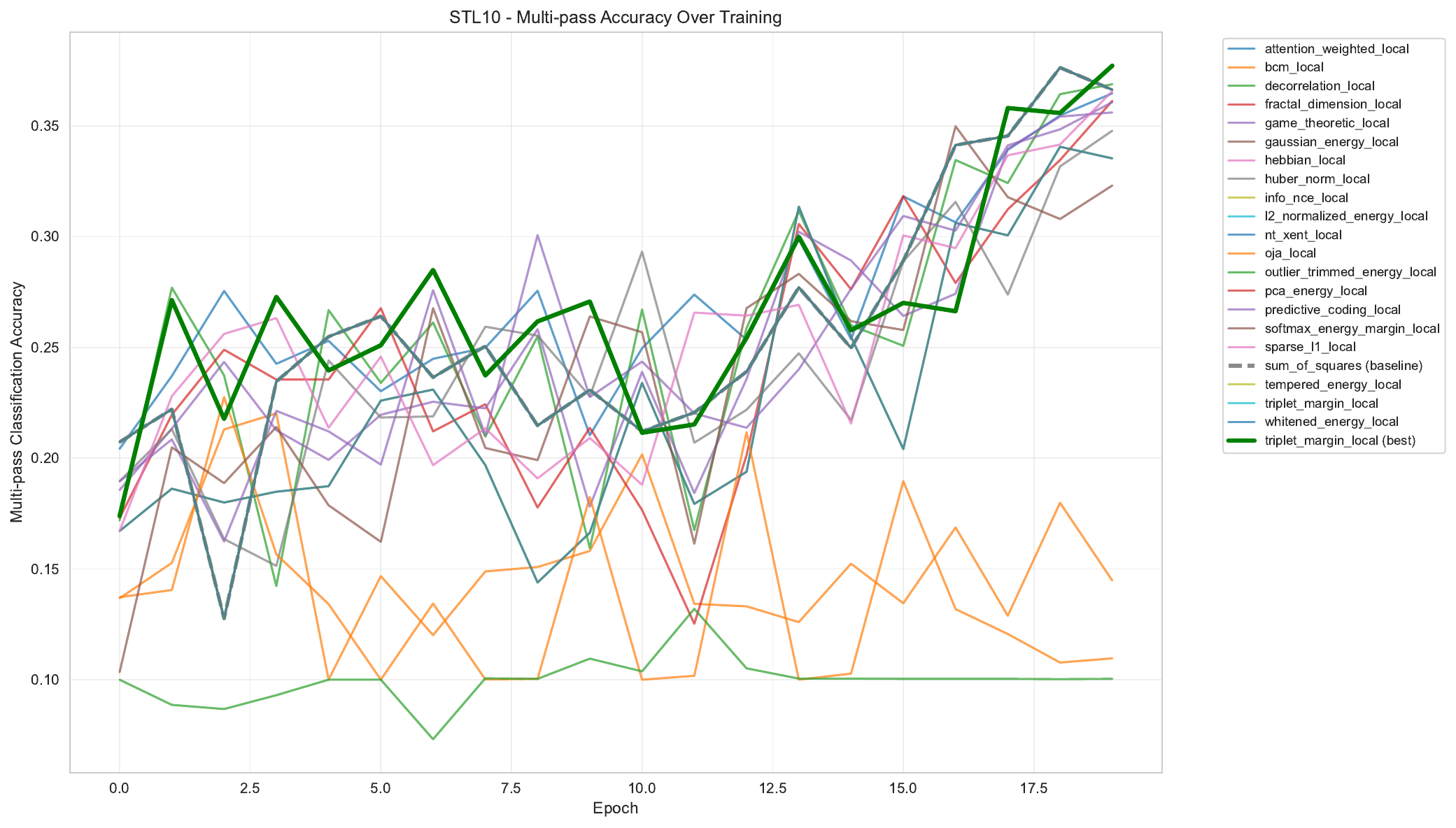}
    \caption{Multi-pass Accuracy progression over training epochs on STL-10. \texttt{triplet\_margin\_local} (green line) demonstrates superior convergence characteristics compared to other functions.}
    \label{fig:stl10_multi}
    \Description{Line graph of multi-pass accuracy for STL-10. The graph shows significant fluctuation. The green line for triplet margin ends at the highest point, indicating the best final accuracy.}
\end{figure}
\end{document}